\documentclass[conference]{IEEEtran}

\usepackage{algorithm}
\usepackage[noend]{algpseudocode}
\usepackage{etoolbox}
\usepackage{enumerate}
\usepackage{enumitem}
\usepackage{amssymb}
\usepackage{amsmath}
\usepackage{mathtools} % amsmath with fixes and additions
\usepackage{booktabs} % commands to create good-looking tables
\usepackage{graphicx}
\usepackage{cite}
\usepackage{url}
\usepackage{hyperref}

\allowdisplaybreaks[4]

\newtheorem{theorem}{Theorem}[section]
\newtheorem{lemma}{Lemma}[section]

\newtheorem{remark}{Remark}

\newenvironment{proof}{{\bf Proof:}}{\hfill\rule{2mm}{2mm}}
\newenvironment{proofsketch}{{\bf Proof Sketch:}}{\hfill\rule{2mm}{2mm}}

\newcommand{\lf}[2]{% \lf{num}{den}
  \ensuremath{\log\left(\frac{#1}{#2}\right)}}

\newcommand\numberthis{\addtocounter{equation}{1}\tag{\theequation}}

\newcommand{\argmin}{\arg\!\min}
\newcommand{\argmax}[1]{\underset{#1}{\operatorname{arg}\,\operatorname{max}}\;}
\newboolean{showcomments}
\setboolean{showcomments}{true}
\usepackage{xcolor}
\newcommand{\jn}[1]{\ifthenelse{\boolean{showcomments}}{{\textbf{(JN says:  #1)}}} {}  }
\newcommand{\nh}[1]{\ifthenelse{\boolean{showcomments}}{{\textbf{(NH says:  #1)}}} {}  }
\newcommand{\yd}[1]{\ifthenelse{\boolean{showcomments}}{{\textbf{(YD says:  #1)}}} {}  }
\newcommand{\review}[1]{\ifthenelse{\boolean{showcomments}}{{\textcolor{red}{(#1)}}} {}  }

\newcommand{\la}{\leftarrow}

\newcommand{\da}{\downarrow}
\newcommand{\prob}[1]{P\left(#1\right)}
\newcommand{\Exp}[1]{\mathbb{E}\left[#1\right]}

\newcommand{\R}{\mathbb{R}}
\newcommand{\N}{\mathbb{N}}

\newcommand{\norm}[1]{\| #1 \|}

\newcommand{\hide}[1]{}
\DeclareGraphicsExtensions{.pdf, .png}

\newboolean{wiopt}
\setboolean{wiopt}{false}
\newcommand{\conftext}[1]{\ifthenelse{\boolean{wiopt}}{{#1}}{}}
\newcommand{\TRtext}[1]{\ifthenelse{\boolean{wiopt}}{}{#1}}

\begin{document}

\title{Unsupervised Crowdsourcing with Accuracy and Cost Guarantees}

\author{\IEEEauthorblockN{Yashvardhan Didwania, Jayakrishnan Nair}
\IEEEauthorblockA{\textit{Department of Electrical Engineering} \\
\textit{IIT Bombay}}
\and
\IEEEauthorblockN{N. Hemachandra}
\IEEEauthorblockA{\textit{Industrial Engineering and Operations Research} \\
\textit{IIT Bombay}}
}
\maketitle

\begin{abstract}
  We consider the problem of cost-optimal utilization of a crowdsourcing platform for binary, unsupervised classification of a collection of items, given a prescribed error threshold. Workers on the crowdsourcing platform are assumed to be divided into multiple classes, based on their skill, experience, and/or past performance. We model each worker class via an unknown confusion matrix, and a (known) price to be paid per label prediction. For this setting, we propose algorithms for acquiring label predictions from workers, and for inferring the true labels of items. We prove that if the number of (unlabeled) items available is large enough, our algorithms satisfy the prescribed error thresholds, incurring a cost that is near-optimal. Finally, we validate our algorithms, and some heuristics inspired by them, through an extensive case study.
\end{abstract}

\section{Introduction}
\label{sec:intro}

% Crowdsourcing is a popular approach. Used for a whole bunch of tasks
% that cannot be done well by algos, and sometimes to create a
% training dataset for ML algorithms. Several kew aspects to consider

%\textcolor{blue}{how far are we exceeding the page limit? (nh)}
Crowdsourcing is an early component of the growing gig economy, and
has been applied in a wide variety of application domains, including
image classification \cite{intro_ic}, image clustering~\cite{intro_ic2}, natural language processing~\cite{intro_nlp}, tagging of fake news and
pseudoscientific content~\cite{intro_news}, graphic design~\cite{intro_gd}, and software development~\cite{intro_swe}. Platforms like Mechanical Turk,
Topcoder, and Designhill allow a requester to recruit workers from
across the globe on-demand to complete prescribed tasks, requiring
varying levels of time, skill and expertise. Increasingly often,
crowdsourcing is also used to create datasets for training machine
learning algorithms.

We consider the task of classifying a collection of
items via crowdsourcing. This might correspond, for example, to the
labeling of images \cite{chang2017revolt}, detecting sarcasm in language \cite{intro_nlp2, intro_nlp}, or labeling online content as incorrect/inappropriate \cite{intro_profanity}. From the
standpoint of the requester who seeks to crowdsource the task(s),
there are several issues/considerations that arise in practice.

%1. Often unsupervised. Experts are expensive.

Firstly, the classification task is often \emph{unsupervised}. In other words, the requester, a.k.a., learning agent, may not have a set of items for which the true label (i.e., the ground truth) is known. Indeed, the very goal of the crowdsourcing activity is often to generate a reliably labeled training dataset that can then be used to train machine learning models that can perform the classification task at scale.

%2. Workers have varying, unknown accuracy levels. Can classify
% workers in terms of certain attributes, but not always clear how
% much a certain attribute would impact accuracy

Secondly, workers on crowdsourcing platforms may have different, and a
priori unknown accuracy levels. (The heterogeneity across workers might
stem from diversity in skill, training, experience, or attention
span.) Moreover, since the accuracy level of a class of workers is a
priori unknown to the requester, ensuring a prescribed end-to-end
accuracy in item labeling (typically by aggregating label
predictions from several workers) is challenging.

% 3. Have to pay more for more more accurate workers. Not always clear
% what is optimal --- larger number of cheaper and less accurate label
% predictions, or fewer more accurate and more expensive predictions

Thirdly, the requester would be concerned about the cost incurred in performing the classification task to the desired level of accuracy. Indeed, the cost of acquiring label predictions from human subjects would influence the number of predictions per item that are collected by the requester. Moreover, many crowdsourcing platforms classify workers into different classes based on their skill level or past performance; this allows for differential pricing of label predictions based on the class the worker belongs to. The presence of multiple worker classes, with different prices, and different (unknown) accuracy levels, makes the optimization of cost non-trivial for the requester.
%\nh{more so if a certain level of accuracy level is required?} 
After all, should the requester acquire a few expensive label predictions per item from more qualified workers, or should she acquire many cheaper label predictions per item from less qualified workers? 
% This work: Poses an optimal classification task taking the above
% aspects into consideration. Give some details.

In this paper, taking the above considerations into account, we consider the problem of optimally utilizing the crowdsourcing platform from the standpoint of the requester. Specifically, we consider the minimization of cost in the unsupervised binary classification of a collection of items, using a multi-class 
%\nh{binary? actually, this phrase `multi-class' of $M$  classes can confuse with `multi-class' in classification setup that generalizes binary classification} 
crowdsourcing platform, given a prescribed error tolerance. We propose novel algorithms for assigning labeling tasks to workers and for estimating true labels; these algorithms are validated via analytical performance guarantees, as well as a case study.

We model each worker class using a latent confusion matrix (as in~\cite{Dawid1979MaximumLE}). 
%\review{DS model includes the confusion matrices and then EM for inferring the true label, this the MLE part that R1 was talking about. We clarify this in Related Lit, but should here as well. Or only confusion matrices}. 
In other words, each worker class is associated with a confusion matrix that specifies the probability that a worker of that class mis-labels an item, as a function of the item's true label. Additionally, each worker class is associated with a price, to be paid per label prediction by a worker of that class. Our algorithms use the powerful tensor-based machinery, pioneered by \cite{JMLR:v15:anandkumar14b}, and developed further by \cite{jordan}, to learn the confusion matrices of each class. These estimates are further used to identify the worker class that 
%\nh{which? `that' is repeated} 
can provide the desired accuracy in label identification at a minimum cost.

Our main contributions are as follows:

%\begin{enumerate}
\noindent {\bf 1.} We formally pose an optimization problem corresponding to the minimization of the cost incurred by a learning agent for binary classification of a collection of items on a multi-class crowdsourcing platform, subject to a prescribed error tolerance (see Section~\ref{sec:setup}). 
    
The key distinction between our model and prior works in the literature is that we associate a confusion matrix with a worker class, rather than each individual worker. While this approach simplifies the task of learning the confusion matrices, it also allows us to define an `optimal' worker class, i.e., the class that provides the desired accuracy at the least cost. This optimal class is task-dependent, and need not be either the most qualified or the least qualified worker class; the identity of the optimal worker class is dictated by the relationship between the (unknown) labeling accuracy of each class with its price.
  
\noindent {\bf 2.} We first consider the special case where the labeling accuracy of each worker class is insensitive to the true item label. In this case, the confusion matrices become symmetric (see Section~\ref{sec:symmetric}). Compared to the general setting (asymmetric confusion matrices), simpler confusion matrix estimation algorithms, with stronger concentration bounds, are available for this special case.
    
We propose a near-optimal algorithm for acquiring label predictions, and for estimating the true item labels in this setting. The algorithm proceeds in two stages---an \emph{explore} stage first identifies the optimal worker class with high probability, and an \emph{exploit} phase collects label predictions using workers from the optimal class identified before, and infers the true labels of the items. 
\hide{In particular, the exploit phase uses the so-called Chernoff stopping criterion, which provably meets the prescribed accuracy threshold, and is also asymptotically (with respect to the accuracy threshold) optimal in the number of label predictions requested.} 
    
\noindent {\bf 3.} Next, we consider the general case where the confusion matrices are asymmetric \conftext{(see Appendix~A in~\cite{FullVersion})}\TRtext{(see Appendix~\ref{sec:asym_appendix})}. We propose and analyse a two-stage algorithm that is structurally similar to the one for the symmetric setting. Confusion matrices are estimated using the spectral tensor methods proposed in \cite{jordan}. 
\hide{As expected, the performance guarantees are somewhat weaker for this algorithm, as compared to the algorithm for symmetric confusion matrices.} 

\noindent {\bf 4.} Motivated by the above algorithms, we propose two heuristics, which attempt to better exploit the information gathered during the explore phase (see Section~\ref{sec:heuristics}). While these approaches are not amenable to analytical performance guarantees, they perform very well in practice.

\noindent {\bf 5.} Finally, we present a case study that validates the performance of the proposed algorithms in practice (see Section~\ref{sec:case_study}). 
%We find that the proposed algorithms outperform naive heuristics, as well as variants of our algorithms that employ an alternative,  commonly used stopping criterion based on confidence bounds. 
%Interestingly, we find that the proposed algorithms (and natural heuristic variants of these) outperform the analytical guarantees we are able to prove. Specifically, we observe that the lower bounds we derive on the number of items, for our performance bounds to hold, are quite loose. In other words, the proposed algorithms perform well even when the number of items to be classified is moderate. 
%\end{enumerate}

%The remainder of this paper is organised as follows. After a brief survey of the related literature below, we state our problem formulation and state some preliminaries in Section~\ref{sec:setup}. We consider the special case of symmetric confusion matrices in Section~\ref{sec:symmetric}, and then the general asymmetric case in Section~\ref{sec:asymmetric}. We present a case study validating our algorithms in Section~\ref{sec:case_study}, and conclude in Section~\ref{sec:conclusion}. 
\conftext{Throughout the paper, references to the appendix (for proofs of our analytical results) point to the appendix in the extended version of this paper~\cite{FullVersion}.}

%\emph{nh says: this para can be avoided; these are covered in listed contributions; just last line is needed }

% This leverages the tensor based machinery pioneered by Anandkumar,
% further developed by Jordan.

%Bulleted list of contributions

%\subsection*{Related Literature}
%\label{sec:related_lit}
%\subsection{Related Literature}
\noindent{\bf Related Literature}

There has been extensive work dedicated to the problem of allocating tasks to different workers and aggregating the labels provided by them to infer the true label of each item \cite{JMLR:v11:raykar10a, Karger2013EfficientCF, liu_nips, pmlr-v32-zhouc14}. The generative model of representing each worker by a latent confusion matrix was proposed by \cite{Dawid1979MaximumLE}, which is quite popular in the crowdsourcing community due to its simplicity \cite{jordan, khetan, ghosh11, dalvi13}. A key assumption behind this generative model is conditional independence of label predictions across the different workers, given the true item label. \cite{Dawid1979MaximumLE} also proposed an inference algorithm based on Expectation-Maximization (EM), and the initialization of this EM algorithm is also an area of interest \cite{jordan, Balakrishnan,Bonald}. The EM algorithm also assumes that all items are equally difficult to classify. 
%
%For the special case of binary classification under symmetric confusion matrices (also referred to as the one-coin model), this limitation has been countered by introducing a difficulty parameter for each item to represent its ambiguity \cite{whitehull, khetan}. The one-coin model is widely used \cite{ghosh11, karger_iterative}.  \cite{pmlr-v51-jaffe16} have attempted to counter the conditional independence among workers by introducing an intermediate layer of latent variables between the true label and worker responses. 
Recently,~\cite{Rangi2018MultiArmedBA} have allowed for different costs for each worker and reduced the problem to a budgeted multi-armed bandit. 

The preceding literature treats each crowdsourcing worker as a distinct individual, and does not group `similar' workers into classes, as is the approach adopted in this paper. 
%We focus on the other notable limitation of statistical independence among the workers on the crowdsourcing platform.  
Indeed, empirical studies on  crowd-labelling platforms suggest that the location, age, cognitive ability, and approval rates  of workers are related to their quality of work; these studies recommend classifying workers along these attributes for appropriate task allocation~\cite{cognitive_acc, kazai, kazai12, loepp}. However, this aspect has received relatively little attention in analytical studies. We seek to fill this gap, by proposing and analysing a model with multiple worker classes, each class being associated with a single (unknown) confusion matrix. A recent paper that takes a different approach towards multiple worker classes is \cite{Shah2018}, which considers binary classification where workers and tasks are of $d$ different types---the prediction accuracy being $p$ if the worker and task type are matched, and 1/2 otherwise. Thus, the crux of the algorithm in \cite{Shah2018} is the clustering of workers into types. In contrast, in the present paper, workers classes are pre-defined, the goal being instead to minimize the cost of reliable labelling.

\begin{remark}
\label{remark:class-based-CF}
It is important to note that in practice, workers within each class may indeed have somewhat different levels of reliability for the specific classification task at hand (in spite of their predictions being priced identically on the platform). Our confusion matrix model should be interpreted as having been averaged over the underlying `reliability distribution' of each class. This is reasonable when i) there is a large number of workers available in each class, and (ii) it is not worthwhile to learn the confusion matrix of each worker separately. In essence, we use the built-in classification of workers as provided by the platform (as, for example, is the case with Amazon MTurk), and treat each class as being composed of a large (and possibly diverse) population of workers who are queried randomly. This \emph{class-centric} approach is different from the \emph{worker-centric} approach that is prevalent in the crowdsourcing literature---the former is not just more efficient from a learning standpoint, but may also be more practical for use in crowdsourcing platforms. Our novelty lies in using the class-centric approach to optimize the cost of labeling the dataset, subject to an accuracy constraint. Such a cost optimization, clearly of practical relevance, has not been addressed in the prior literature.
\end{remark}

%attempt to advance the body of work in this regard by assuming that we have heterogeneous worker classes , each of which represent a homogeneous group of workers with similar demographics and abilities. 

\hide{
From an analytical perspective, \cite{pmlr-v51-jaffe16} have attempted to address the 
d there have been some studies  \yd{rephrase..'some studies'} to identify sets of dependent workers based on their responses in an unsupervised setting \cite{jaffe, donmez, platnios}. However , there has been relatively fewer work to address this limitation to the best our knowledge. We attempt to \yd{some good phrase for 'providing insight' or 'advance the work'} by introducing heterogenity in worker classes
 }

%Theres been lot of work  in crowdsourcing , we re going to cite only a few. Beginngiwth DS and different ways on which EM was improved and finally by jordan 
%  Several algos for true label aggregation from Single parameter workers corresponding to symmetric conf martrices, have been proposed [][][]
% These are all one -shot
% Afterwards a difficulty param for each task was introduced to make the case for better adaptve task alloc algorithms  that outperfrom non -adaptive algos. [][][]

%with asymmetric there has also been some work [][][]
% Althouth the work has been extended to $\Kappa$-ary classification of items, there has been little attention in analytical studies towards different classes of workers available on crowdsourcing platforms. Empirical studies show that workers hvae differnet backgrounds and qualifications and these significantly impact their quality of labels. In particualr, so and so show that only tweaking the monetary rewards and HITs attract substantially different workers pointing to the heterogenity in workers.

% We work on heterogenity of workers while most analytical studies have done heterogenity in tasks

\hide{

The Dawid-Skene model has been extensively used in crowdsourcing literature to  model the ability of crowdsourcing workers for multi-class labelling . \cite{Dawid1979MaximumLE} infer the true labels of tasks with an algorithm based on Expectation-Maximization. Depeding on the initialization of the EM algorithm, it can converge to a local optimum and hence performance guarantees were not available. Using spectral methods developed by \cite{JMLR:v15:anandkumar14b} to initialize the EM algorithm,  \cite{jordan} provided guarantees on the convergence of the estimates fo the confusion matrices to their true values. In the context of binary labelling tasks \cite{NIPS2011_c667d53a} introduced the \emph{spammer-hammer} model where a worker can always assign correct labels (hammer) or randomly assign labels to items (spammer). This was later extended to the $k$-ary classification tasks by formulating it as $k-1$ binary tasks \cite{Karger2013EfficientCF}.  In our work, we represent this model as symmetric confusion matrices in Section \ref{sec:symmetric}.
 For the binary tasks,  \cite{NIPS2011_c667d53a} provide one-shot task allocation and iterative inference algorithms optimal in the number of workers needed per item to achieve a target classification accuracy. This non-adaptive algorithm asymptotically achieves the fundamental minimax error rate , hence adaptivity offers only marginal gains. \yd{rephrase ?} . 
With the observation that this is due to all  tasks being equally difficult  , \cite{NIPS2016_03e7ef47} introduced a difficulty parameter (originally proposed\cite{NIPS2009_f899139d}) for every item and proposed adaptive algorithms that outperforms non-adaptive schemes\yd{Can add more details about task alloc using bipartite graphs .}. \par More recently, \cite{Rangi2018MultiArmedBA} allowed for different costs for each worker and reduced the problem to a Budgeted Multi-Armed Bandit. They further show that the $B-KUBE$ algorithm has logarithmic regret in budget and empirically outperforms other MAB algorithms in this setup. In our work, since the objective is to..\yd{say that we are optimizing a function of the confusion matrix, and therefore ones with highest conf are not best choice as in MAB.}
\yd{Also want to clearly state that all of the above binary tasks are for symmetric case with a single parameter, in which case knowledge of the priors is not required. But for the asymmetric case I did not find any except \cite{jordan}}
}

\section{Model and Preliminaries}\label{sec:setup}

In this section, we describe our model for unsupervised labelling of items using crowdsourced label predictions, and establish the benchmark that we evaluate our algorithms against.

We begin by defining some notation. 
%We use the abbreviations UCB and LCB for \emph{upper confidence bound} and \emph{lower confidence bound}, respectively. 
For $n \in \N,$ $[n]:=\{1,2,\cdots,n\}.$ The indicator function $\mathbf{I}(z)$ equals~1 if condition~$z$ is true, and 0 otherwise. For $x \in \R,$ $\mathrm{sign}(x):= \mathbf{I}(x \geq 0) - \mathbf{I}(x < 0).$ Finally, let $d(\cdot , \cdot)$ denote the KL-divergence between two Bernoulli distributions (also commonly referred to as the binary relative entropy), i.e., for $p,q \in (0,1),$ 
\begin{equation*}
    d(p,q) := p \log \left(\frac{p}{q}\right) + (1-p) \left(\frac{1-p}{1-q}\right).
\end{equation*} 

\subsection{Problem formulation}

We consider a binary classification task, where we are given~$N$ items, where the true label $\ell_j$ of item $j \in [N]$ lies in $\{0,1\}.$ We further assume that $(\ell_j,\ j \in [N])$ is an i.i.d. Bernoulli random vector with $\prob{\ell_j = i} = w_i$ for $i \in \{0,1\}.$ In other words, the true labels of the different items are independent, taking value~0 with probability~$w_0,$ and~1 with probability~$w_1 = 1-w_0.$ We refer to $w = (w_0,w_1)$ as the \emph{prior} on the true labels. We note that both the true labels $(\ell_j,\ j \in [N])$ as well as the prior~$w$ are a priori unknown to the learning agent. The goal of the learning agent is in turn to accurately estimate the true label of each item with high probability. This estimation is performed using label predictions on a multi-class crowdsourcing platform, which is described next.

%\nh{i think we need to use either one of $[n]$ or $[N]$ in above} JK: Addressed

%\nh{also, let's define $[n]$ as $[n] : = \{1, \cdots, n\}$} JK: Addressed

The crowdsourcing platform consists of $M \geq 3$ classes of workers,\footnote{We assume $M \geq 3$ because this is required by the spectral machinery for confusion matrix estimation that we leverage. The case~$M=2$ can be shown to be fundamentally ill posed in the unsupervised setting.} where each worker class may be defined by certain qualifications (like academic background, age, gender, nationality, etc.) and/or past performance on the platform.\footnote{For example, on Amazon MTurk, workers can be classified based on number of tasks completed and their approval rates.} A worker of class~$k,$ $k \in [M],$ charges a price~$p_k$ per label prediction. We model worker reliability/performance 
%\review{use "performance" to avoid any confusion? also can say we ignore dependencies on task design/complexity, order of tasks, etc.} 
using the latent confusion matrix model proposed by \cite{Dawid1979MaximumLE}: The labels assigned by different workers to an item~$j$ are conditionally independent, given the true label $\ell_j$. Moreover, the labelling accuracy of workers of class~$k$ is characterized by a latent confusion matrix
\begin{equation*}
    \textbf{C}_k :=  \begin{bmatrix} c_k(0) & 1 - c_k(1)\\ 
    1 - c_k(0) & c_k(1)
    \end{bmatrix},
\end{equation*}
where $c_k(i) \in (0,1)$ is the probability of correctly labelling an item with true label~$i \in \{0,1\},$ by any worker of class~$k$. We further assume that $c_k(0), c_k(1) > \frac{1}{2}$ for all classes~$k,$ i.e., given any item, any worker is more likely to predict the true label correctly than a random guess. We also use the notation $c_k(i,j)$ to refer to the entry corresponding to true label~(column)~$i$ and predicted label~(row)~$j$ in the confusion matrix $\textbf{C}_k$. For example, $c_k(0,1) =1-c_k(0).$ 

\hide{we will refer to the $i^{th}$ diagonal entry by $c_k(i)$ and}

To summarize, each worker of class~$k$ is characterized by a price~$p_k$ to be paid for each label requested, and confusion matrix~$\textbf{C}_k$. Note that the confusion matrices are also a priori unknown to the learning agent. Also, recall that as noted in Remark~\ref{remark:class-based-CF}, $\textbf{C}_k$ should be interpreted as having been averaged over the underlying `reliability distribution' of class~$k,$ with any labelling request from a class~$k$ worker being routed to a randomly selected worker from this class.

For the above model, the goal of the learning agent is to predict the true labels of the~$N$ items, such that each item's true label is identified correctly with probability at least $1-\alpha,$ where $\alpha \in (0,1)$ is a prescribed error tolerance, while minimizing the cost of acquiring label predictions on the crowdsourcing platform. (Our cost benchmark is formalized below.) Finally, it is important to emphasize that we consider an \emph{unsupervised} setting, i.e., there is no labeled training dataset (a collection of items whose true labels are a priori known) that the agent can use to learn the confusion matrices.

%\nh{apart from `classes', we can say `groups' of workers. Later, we can say that our computational setup considers 3 groups of workers}

\subsection{Optimal worker class}
\label{sec:benchmark}

In order to pose the problem of cost-optimal label prediction subject to an accuracy guarantee, we now define the cost-optimal worker class, denoted by~$k^*.$ We begin by bounding from below the cost of estimating the true label of a single item accurately with probability at least~$1-\alpha,$ using label predictions from workers of class~$k$, \emph{assuming no prior knowledge of the confusion matrices}. An alternative lower bound, that assumes perfect knowledge of the confusion matrices, is presented in Section~\ref{sec:heuristics}.

%\footnote{An alternative lower bound, that assumes prior knowledge of the confusion matrices, will be presented in Section~\ref{sec:heuristics}.}
%\jn{How about using $d(p,q)$ instead, and just refer to it as binary relative entropy? Also fine with $kl(p,q),$ as suggested by NH.\yd{addressed. keeping $KL()$}}

\begin{lemma}\label{lemma:infoT}
Consider an item $j$ with (unknown) true label $\ell_j$, and a worker class~$k.$ For $\alpha\in (0,1),$ consider any algorithm that identifies the true label $\ell_j$ with probability at least $1-\alpha,$ using only label predictions on item~$j$ from worker class~$k,$ with no prior knowledge of~$\textbf{C}_k.$ Then the number of label predictions~$\tau_k(\ell_j)$ collected by the algorithm satisfies
\begin{equation*}
    \Exp{\tau_k(\ell_j)} \geq \frac{1}{d(c_k(\ell_j), 0.5)}\log\left(\frac{1}{2.4\alpha}\right).
\end{equation*}

\end{lemma} 

Lemma~\ref{lemma:infoT} provides an information theoretic lower bound on the number of label predictions (or queries) needed from class~$k$ workers in order to identify the true label of item~$j$ with probability at least~$1-\alpha.$ As expected, the lower bound is dictated by how close~$c_k(\ell_j)$ is to 1/2; the further away it is from 1/2, i.e., the more accurate class~$k$ workers are at predicting the true label~$\ell_j,$ the smaller is the lower bound on the average number of queries. Lemma~\ref{lemma:infoT} is proved by mapping the problem of true label identification to a certain multi-armed bandit (MAB) problem and then invoking Theorem 6 of \cite{JMLR:v17:kaufman16a}; \conftext{see Appendix~C in~\cite{FullVersion}}\TRtext{see Appendix~\ref{sec:appendix1}}.

Lemma~\ref{lemma:infoT} implies that the expected number of class~$k$ queries required to meet the prescribed accuracy guarantee for a typical item is at least 
\begin{equation}
\label{eq:lower_bound_with_prior}
  \log\left(\frac{1}{2.4\alpha}\right)\left(
    \frac{w_0}{d(c_k(0), 0.5)} + \frac{w_1}{d(c_k(1), 0.5)} \right).  
\end{equation}
Accordingly, we define the cost-optimal worker class~$k^*$ as follows:
\begin{align}
    k^* &= \argmin_{k \in [M]}  \sup_w  \biggl[p_k \log\left(\frac{1}{2.4\alpha}\right) \nonumber \\
    & \qquad \qquad \left(
    \frac{w_0}{d(c_k(0), 0.5)} + \frac{w_1}{d(c_k(1), 0.5)} \right) \biggr] \nonumber\\
    &= \argmin_{k \in [M]} \biggl[p_k\max\left(\frac{1}{d(c_k(0), 0.5)},\frac{1}{d(c_k(1), 0.5)} \right) \biggr]
    \label{eq:optclass}
\end{align} 
Note that the above definition of the optimal worker class is dependent only on the confusion matrix and the price per label corresponding to each worker class. Specifically, it does not depend on the prior~$w;$ it considers instead a `worst case' of the lower bound \eqref{eq:lower_bound_with_prior} over all priors.\footnote{Alternative formulations, based on either estimating the prior, or simply assuming one, are also possible using the same machinery.} However, in the special case where the confusion matrix is symmetric (this case is addressed in Section~\ref{sec:symmetric}), the information theoretic lower bound in~\eqref{eq:lower_bound_with_prior} is insensitive to the prior, making the above `worst case' operation redundant. Note also that the optimal worker class also does not depend on the error threshold~$\alpha.$ For simplicity, we assume that the minimizer~$k^*$ in \eqref{eq:optclass} is unique, i.e., there is a unique optimal worker class.\footnote{This assumption is made purely to simplify the presentation of our performance guarantees and their proofs; the extension to the case of multiple optimal worker classes is trivial.}

Finally, we define sub-optimality gaps~$\Delta_{k}$ for $k\in[M]$ as follows. Towards this, we first define, for $k \in [M],$ $$s_k := p_k\max\left(\frac{1}{d(c_k(0), 0.5)},\frac{1}{d(c_k(1), 0.5)} \right).$$ For $k \in [M]\setminus \{k^*\},$ $\Delta_k := s_k - s_{k^*},$ 
%\begin{align*}
%    \Delta_k &:= p_k\max\left(\frac{1}{d(c_k(0), %0.5)},\frac{1}{d(c_k(1), 0.5)} \right) \\ &\qquad %\qquad - p_{k^*}\max\left(\frac{1}{d(c_{k^*}(0), %0.5)},\frac{1}{d(c_{k^*}(1), 0.5)} \right)
%\end{align*}
and $\Delta_{k^*} = \Delta_{min}  := \min_{k \in [M]\setminus\{k^*\}}\Delta_k.$
%\yd{Prof. JK, please see if this can be made compact with $s_k$} %JK: Looks fine to me.
\hide{
It now follows from Lemma \ref{lemma:infoT} that in order to bound the probability of error (in label prediction for a single item) from above by~$\alpha,$ it suffices to obtain~$s_k$ label predictions from class~$k$ workers, where
\begin{equation}
    s_k = \frac{1}{d(\min\{c_k(0), c_k(1)\}, 0.5)}\log\left(\frac{1}{2.4\alpha}\right)
\label{eq:no_labels}
\end{equation}
This is of course assuming that we also have an inference algorithm that closely match the lower bounds of Lemma \ref{lemma:infoT}.
The total cost of acquiring these label predictions would then be $p_k s_k,$ for the worker class $k$ under consideration. 
Thus, we define the cost-optimal worker class~$k^*$ as follows: \begin{align}
    k^* &= \argmin_{k \in [M]}  \frac{p_k}{d(\min\{c^{1}_k, c^{2}_k\}, 0.5)}\log\left(\frac{1}{2.4\alpha}\right) \nonumber \\
    &= \argmin_{k \in [M]} \max \left\{ \frac{p_k}{d(c_k(0), 0.5)}, \frac{p_k}{d(c_k(1), 0.5)}  \right\}
    \label{eq:optclass}
\end{align} 
} % end hide

\hide{
Our cost benchmark per item, given the error tolerance~$\alpha,$ is then defined as 
\begin{equation*}
p_{k^*} s_{k^*} = p_{k^*} \max \left\{ \frac{1}{d(c_k(0), 0.5)}, \frac{1}{d(c_k(1), 0.5)}  \right\}.
\end{equation*}
Finally, we define the sub-optimality gap~$\Delta_{\min}$ as follows:
\begin{equation*}
    \Delta_{\min} = \min_{k \in [M]\setminus\{k^*\}} p_k s_k - p_{k^*}s_{k^*}
\end{equation*}
} % end hide

In the following sections, we evaluate our learning algorithms 
in terms of 
 \begin{enumerate}
    \item the probability that the estimated optimal worker class~$\hat{k}$ equals $k^*$ (note that our algorithms \emph{do not} know the confusion matrices a priori), 
    %\nh{and also don't use/know prior class probabilities $(w_1, w_2)$? and also don't use labels of the training data; in fact, the algorithm outputs labels},
    and 
    %\nh{i think we should emphasise \emph{we don't know} or \emph{do not} or just \emph{not -- this is better}. not sure of  emphasising `know'}
    \item the expected number of queries requested per item (benchmarked against the lower bound of Lemma~\ref{lemma:infoT} applied to the worker class~$k^*$).
    %a high probability upper bound on the difference between the cost incurred per item by the algorithm and the benchmark $p_{k^*} s_{k^*}.$  
    %\yd{Mention that we denote this as $B$ ?} \jn{Not needed, IMO. Unless we are including the bound on $B$ here.}

%    additional cost per item needed per item, $B = p_{\hat{k}}\hat{s}_{\hat{k}} - p_{k^*}s_{k^*}$
\end{enumerate}

Note that the cost benchmark noted above (obtained by applying  Lemma~\ref{lemma:infoT} to the worker class~$k^*$) is somewhat weak, since it applies to algorithms that perform label assignment without prior knowledge of the confusion matrices. However, note that identifying $k^*$ in the first place involves estimating the confusion matrices. This suggests the possibility of exploiting these confusion matrix estimates to lower the labelling cost per item further below the above mentioned benchmark. We propose heuristic approaches that do this in Section~\ref{sec:heuristics}; see also Remark~\ref{rem:DT_versus_heuristic}.

In Section~\ref{sec:symmetric}, we consider the special case where the reliability/performance of the worker classes does not depend on the underlying true label of the item, i.e., $c_k(0) = c_k(1).$ \hide{In this case, the confusion matrices are symmetric, and this structure can be exploited to obtain stronger performance guarantees, as compared to the general case (with asymmetric confusion matrices), which, due to space constraints, is addressed in Appendix~\ref{sec:asym_appendix}.} The general case (with asymmetric confusion matrices), due to space constraints, is addressed in \conftext{Appendix~A in~\cite{FullVersion}}\TRtext{Appendix~\ref{sec:asym_appendix}}.
\section{Symmetric Confusion Matrices}\label{sec:symmetric}

In this section, we consider a special case of our model where the accuracy of workers is insensitive to the true item labels. This corresponds to a confusion matrix where the diagonal elements are equal, so that each confusion matrix $\textbf{C}_k$ is parameterized by a single parameter $c_k = c_k(0) = c_k(1).$ 
%(In this case, note that the prior distribution~$w$ of the true labels become irrelevant.) 
This model, wherein each worker provides accurate labels with a certain probability, is also commonly referred to as the one-coin model in the crowdsourcing community; examples of papers that adopt this model include \cite{karger_iterative, Rangi2018MultiArmedBA, khetan}. For this model, we use the results for the ``one-coin model" in  \cite{jordan} to estimate, and derive concentration inequalities on, the confusion matrices of all classes.

The proposed algorithm for the case of symmetric confusion matrices, which we refer to as Sym--IMCW (IMCW stands for {\bf I}nference using {\bf M}ulti-{\bf C}lass {\bf W}orkers) is stated formally as Algorithm~\ref{alg:sym}. This algorithm proceeds in two phases: an \emph{explore} phase, followed by an \emph{exploit} phase. The goal of the explore phase is to estimate the confusion matrices of all classes, and to identify, with high probability, the optimal worker class~$k^*$. Next, in the exploit phase, true labels of all items are estimated using only predictions from the estimated optimal worker class. Interestingly, both phases are structurally similar to (different) MAB algorithms. The explore phase is akin to a \emph{fixed budget} MAB algorithm (where arms correspond to worker classes). On the other hand, the exploit phase, which defines a stopping time criterion to cease the collection of label predictions for each item, is akin to a \emph{fixed confidence} MAB algorithm. In the following, we provide a detailed description of each phase.

\subsection{Explore Phase: Estimating Optimal Worker Class}

The explore phase, defined over lines 3--7 in Algorithm~\ref{alg:sym}, proceeds as follows. First, a single label prediction is acquired for all the given $N$ items, from each worker class (i.e., $M$ label predictions each for~$N$ items). That the true labels are independent and identically distributed across our items is then exploited to estimate the confusion matrices using the spectral techniques developed in~\cite{jordan}. For each worker class~$k,$ the estimated confusion matrix parameter $\hat{c}_k$ is then used to estimate $s_k$ as follows: $$\hat{s}_k:= \frac{p_k}{d( \max(\hat{c}_k , ~ 0.5), 0.5)}$$ The optimal arm is then estimated as $\hat{k} = \argmin_k{\hat{s}_k}.$ 

\begin{algorithm}[t]
\caption{Symmetric-IMCW }
\begin{algorithmic}[1]
    \State \textbf{Input}: prices $\{p_k : k \in [M]\}$; error threshold $\alpha$; number of items $N$;  items $\{j : j \in [N]\}$ 
    \State $\cdots \cdots \cdots \cdots \cdots \cdots \cdots\cdots\cdots\cdots$ \Comment{Explore Phase}
    
    \For{$j = 1, \dots, N$}     
        \ForAll{$k \in [M]$}
        \State Collect 1 predicted label on item $j$ \hide{from worker $k$}  at price $p_k$
        \EndFor
    \EndFor
    \State Run Steps (1)-(2) of Algorithm 2 of \cite{jordan} to obtain $\hat{c}_k$  $\forall\ k \in [M]$
    
    \State Set $\hat{k} = \argmin_{k \in [M] } \frac{p_k}{d( \max(\hat{c}_k , ~ 0.5), 0.5)}$
    \State $\cdots \cdots \cdots \cdots \cdots \cdots\cdots\cdots\cdots\cdots$ \Comment{Exploit Phase}    \ForAll{$j \in [N]$ }       
        \State Assign \textit{final label} $\hat{\ell}_j =$  DirectionTest($\hat{k}$, $j$, $\alpha$)
    \EndFor
\end{algorithmic}
\label{alg:sym}
\end{algorithm}

The estimation of the confusion matrices is based on part of Algorithm 2 of \cite{jordan}, for the ``one-coin model" in crowdsourcing. For completeness, we summarize the relevant steps here. 

For each pair of worker classes $a$ and $b$, define the second order quantity $N_{ab}$ as  

\begin{equation*}
    N_{ab} = \frac{1}{2}\left( \frac{\sum^N_{j=1}{\mathbf{I}{(z_{aj}=z_{bj})}}}{N} - \frac{1}{2} \right), 
\end{equation*}

where $z_{kj}$ is the label assigned by the class~$k$ worker to item~$j$. Note that $N_{ab}$ captures the agreement in label predictions between the (workers picked from) classes~$a$ and~$b.$ %\nh{$\mathbf{I}$ defined? seems indicator fn?} JK: I have, now, in Section 2.
Next, for every worker class $k,$ define $(a_k,b_k)$, and compute the estimate~$\hat{c}_k,$ as follows:
\begin{align*}
    (a_k, b_k) &= \argmax{(a,b):~a\neq b\neq k } \mid N_{ab} \mid \\
    \hat{c}_k  &= \frac{1}{2} + \mathrm{sign}(N_{ka_1})\sqrt{\frac{N_{ka_k}N_{kb_k}}{N_{a_k b_k}}}.
\end{align*}
%\jn{Is that $a_1$ in the $N_{ka_1}$ above?\yd{Yes}} \nh{to check again} 
The final step is to check whether $\frac{1}{M}\sum^M_{k=1}\hat{c}_k \geq \frac{1}{2}$, if not then we update $\hat{c}_k \la 1-\hat{c}_k ~\forall~k$.\footnote{Note that in the unsupervised setting under consideration, $(\hat{c}_k,\ k \in [M])$ is just as consistent with the data as $(1-\hat{c}_k,\ k \in [M]).$ The ambiguity is resolved using the assumption that $c_k > \frac{1}{2}$ for all classes~$k.$}

Next, we describe the exploit phase of Sym--IMCW.

%Lemma~13 of~\cite{jordan} provides a PAC bound on this estimate of the following form: the estimates are $\epsilon$-accurate with probability $\geq 1-\delta,$ if $N$ is large enough, where the values of $\epsilon,$ $\delta,$ and $N$ are constrained in terms of the problem parameters (specifically, bounds on the confusion matrices). In practice we found this PAC bound to be quite loose and instead prefer to use a plug-in estimate. This makes Sym--IMCW and our  following analysis robust to the loose guarantees of \cite{jordan} although we borrow from their algorithm. We also prove separate performance guarantees on our estimator in Theorem \ref{thm:sym}. \yd{Expand on advantages of using plug-in estimate rather than $\epsilon,$-dependent}

%as shown on Line 7 of the algorithm.
%Since we require ~$\epsilon,$ $\delta$  to be algorithm-computable, 

%The dependence on ~$\epsilon,$ and $\delta$ is not completely ignored however. In order to present the guarantees of Theorem \ref{thm:sym}, we parameterize them in terms of a hyperparameter~$\gamma\in\left(0,\frac{1}{2}\right)$ which is made clear in the proof of Theorem \ref{thm:sym} in Appendix \ref{sec:appendix1}

\subsection{Exploit Phase: Fixed Confidence Label Prediction}

Having estimated the optimal worker class in the explore phase, in the exploit phase, we estimate the true item labels. For this, we use our assumption that $c_k > 1/2$ for all~$k,$ reducing the problem of assigning a final label to each item to that of deciding the direction of bias of a biased coin. Accordingly, Algorithm~\ref{alg:dtest}, which assigns a final label to each item (see line~10 of Algorithm~\ref{alg:sym}), has been named \emph{DirectionTest}.

The DirectionTest algorithm (see Algorithm~\ref{alg:dtest}) works as follows. For the given item, we acquire label predictions sequentially, using workers from class $\hat{k}.$ A certain stopping criterion (see line~4 of Algorithm~\ref{alg:dtest}) determines when to stop collecting predictions, and a certain decision rule (see line~8 of Algorithm~\ref{alg:dtest}) determines the final label to be assigned.
%These predictions are then used to estimate the true label of each item via some decision rule.
The algorithm is based on the following observation: For an item~$j$,  $c_{\hat{k}}(\ell_j, 1) > 1/2 \iff \{\ell_j =1\}$. Thus, to identify the true label with probability~$\geq 1-\alpha$, it suffices to determine, with probability~$\geq 1-\alpha$, which of the following holds: $c_{\hat{k}}(\ell_j, 1)>1/2,$ or $c_{\hat{k}}(\ell_j, 1) < 1/2$. 

We model the above determination as a two-armed MAB problem. In this MAB problem, the rewards from arm~1 correspond to the successive label predictions sought for the item consider consideration; this implies arm~1 has a $\mathrm{Bernoulli}(c_{\hat{k}}(\ell_j, 1))$ reward distribution. Arm~2 is a virtual arm, and has a known deterministic reward of~$1/2$; thus arm~2 is never actually `pulled.' For this MAB instance, the condition that arm~1 is optimal (i.e., it has a higher mean reward) is equivalent to the condition $c_{\hat{k}}(\ell_j, 1)>1/2,$ which in turn is equivalent to the true label being~1.

%Our objective is now to compute the better arm. 
The above equivalence allows us to invoke the rich literature on the fixed confidence best arm identification problem for MABs. 
%Applying the MAB machinery, we estimate whether $\mathbf{I}(c_{\hat{k}}(\ell_j, 1)\geq 0.5) $ and consequently estimate $\hat{\ell}_j$ for each item.
Specifically, we rely on the Chernoff stopping rule (first applied to MAB problems in \cite{pmlr-v49-garivier16a}  and \cite{JMLR:v17:kaufman16a}). This boils down to maintaining the running log likelihood ratio between the two hypotheses, and to stop when it exceeds the threshold~$\beta(t, \alpha) = \log\left(\frac{2t}{\alpha}\right);$ here, $t$ denotes the number of queries made so far. At that point, the optimal arm is estimated to be 1 (or equivalently, the true label for the item is estimated to be~1) if~$\hat{c}>1/2.$
%and the threshold $\beta(t, \alpha)$ developed in \cite{pmlr-v49-garivier16a}  and \cite{JMLR:v17:kaufman16a}. We maintain a running likelihood of obtaining predicted label was~$1$ in $\hat{c}$ and stop collecting labels when $d(\hat{c}, 0.5) \geq \frac{\beta(t, \alpha)}{t}$. 
This stopping criterion not only ensures that the labelling accuracy of each item is at least $1-\alpha,$ but also results in an asymptotically optimal query complexity, matching the information theoretic lower bound in Lemma~\ref{lemma:infoT} as $\alpha \da 0.$ This is formalized in Lemma~\ref{lemma:dtest} below.
%Theorem \ref{lemma:dtest} also proves an asymptotic bound for the stopping time of DirectionTest. As $\alpha\ra 0$, the stopping time approaches the lower bound in Lemma \ref{lemma:infoT} in order.

\begin{algorithm}[t]
\caption{DirectionTest}
\begin{algorithmic}[1]
    \State \textbf{Input}: Worker class $k$, Item $j$, Error tolerance $\alpha$
    \State Initialize $t=0, \hat{c}=0$
    \State Set $\beta(t, \alpha) = \lf{2t}{\alpha}$
    \While{ $t\leq 1$ OR $t\ d(\hat{c}, 0.5) \leq \beta(t, \alpha)$ }
    \State $t \la t+1$% and update $\beta$ accordingly
    \State Collect a label prediction $z_t\in \{0,1\}$ on item $j$ from worker class $k$
    \State Update $\hat{c} = \frac{1}{t} \sum_{t} z_i $
    \EndWhile
    \State \textbf{return} $ \mathbf{I}(\hat{c} > 0.5)$ %(and in case of a tie, arbitrarily)  
\end{algorithmic}
\label{alg:dtest}
\end{algorithm}

\begin{lemma}\label{lemma:dtest}
Given a worker class~$k \in [M]$, an item~$j$ with an unknown true label~$\ell_j\in \{0,1\},$ and an error tolerance $\alpha\in (0,1)$, the output $\hat{\ell}_j = \mathrm{DirectionTest}(k, j, \alpha)$ of Algorithm \ref{alg:dtest} satisfies
%running with loop counter $t$ and the stopping threshold $\beta(t, \alpha)$  as follows 
%\begin{equation}\label{eq:beta}
%    \beta(t,\alpha) = %\log\left(\frac{2t}{\alpha}\right)
%\end{equation}, ensures that $\hat{\ell}_j = DirectionTest(k, j, \alpha)$ satisfies
$\prob{\hat{\ell}_j \neq \ell_j }\leq \alpha$ and \begin{equation}\label{eq:tau_as}
     P\left(\limsup_{\alpha\da 0} \frac{\tau_{k}}{\log(1/\alpha)}\leq \frac{1}{d(c_k(\ell_j), 0.5)}\right) = 1
\end{equation}
where $\tau_{k} = \inf\{t\in \N : t\ d(\hat{c}, 0.5) > \beta(t, \alpha) \}$ is the stopping time of Algorithm~\ref{alg:dtest}.
%\review{This $\tau$ cannot be changed according to convention above}
\end{lemma}

\begin{proofsketch}
Theorem 10 of \cite{pmlr-v49-garivier16a} guarantees that for any sampling strategy, the Chernoff stopping rule % with threshold as given in ~\eqref{eq:beta} ensures 
satisfies an $\alpha$-PAC guarantee. %that the better of the two arms is selected at the end with probability $\geq 1-\alpha$.\\
For the claim on asymptotic optimality, we invoke Proposition 13 of \cite{pmlr-v49-garivier16a}. %For our set of arms, $w^*_1 = 1$ and $w^*_2 = 0$. Consequently we only sample the first arm, ensuring $N_1(t) = t$ which satisfies all the conditions of Proposition 13 that guarantees that ~\eqref{eq:tau_as} holds for  all $\alpha\in(0,1).$ In other words, the stopping time algorithm achieves the information theoretical lower bound almost surely as $\alpha$ goes down to 0
\end{proofsketch}

\begin{remark}
\label{rem:DT_versus_heuristic}
It is important to note that our exploit phase algorithm, beyond the input~$\hat{k},$ does not use the confusion matrix estimates generated in the explore phase. This was done to enable meaningful analytical guarantees. In practice, confidence intervals on the confusion matrix estimates from~\cite{jordan} tend to be \emph{very} loose, and baking these intervals into a sound stopping time algorithm would result in considerable over-querying in the exploit phase. A heuristic approach would be to simply ignore the uncertainty in the confusion matrix estimation, and simply apply a stopping time algorithm that is optimal in the (hypothetical) setting wherein the confusion matrices are known a priori. Two such approaches are described in Section~\ref{sec:heuristics}; they cannot be justified analytically, but perform very well in practice (see Section~\ref{sec:case_study}).
\end{remark}

\subsection{Performance Guarantee}

%Finally, we state a formal performance guarantee for the algorithm Sym--IMCW. 

\begin{theorem} 
\label{thm:sym}
Under the Sym--IMCW algorithm, for each item~$j \in [N],$ $\prob{\hat{\ell}_j \neq \ell_j} \leq \alpha.$ Moreover, for some $\gamma \in (0,1/2),$ if $N \geq N_0(\gamma),$ where $N_0$ is a constant that depends on the instance and the hyperparameter~$\gamma,$
\begin{equation*}
     \prob{ \hat{k} \neq k^*} \leq  
    M^2\exp\left(-\frac{N^{1-2\gamma} }{2}\right).
\end{equation*}
Finally, denoting the number of label predictions acquired for item~$j$ in the exploit phase by $\tau_{\hat{k}},$ 
\begin{equation*}
     P\left(\limsup_{\alpha\da 0} \frac{\tau_{\hat{k}}}{\log(1/\alpha)}\leq \frac{1}{d(c_{\hat{k}}(\ell_j), 0.5)}\right) = 1.
\end{equation*}
%\review{The denotion of $\tau$ in terms of unknown $\ell_j$ is ok here? we are not implying that we know $\ell_j$ }
\end{theorem}

The (somewhat cumbersome) expression for $N_0(\gamma)$ is provided in \TRtext{Appendix~\ref{sec:appendix1}}\conftext{Appendix~C in~\cite{FullVersion}} , which also contains the proof of Theorem~\ref{thm:sym}. The main takeaways from Theorem~\ref{thm:sym} are as follows:

%\begin{itemize}
$\bullet$ Sym--IMCW meets the prescribed accuracy guarantee, i.e., the true label of each item is identified with probability at least~$1-\alpha.$

$\bullet$ If $N$ is large enough, 
%(specifically, larger than an instance-dependent constant~$\max(K_1,K_2,K_3)$), \nh{need to avoid thees $K_i$s? now} 
the optimal worker class is identified in the explore phase with high probability. 
%In particular, the probability of mis-identification is at most $M^2\exp\left(-\frac{N^{1-2\gamma} }{2}\right).$ 

$\bullet$ For the estimated optimal worker class~$\hat{k},$ the query complexity in the exploit phase is asymptotically (as $\alpha \da 0$) optimal. This means that Sym--IMCW is, with high probability, nearly cost-optimal (admittedly, relative to the `weak' benchmark indicated by Lemma~\ref{lemma:infoT}). Since the explore phase only requires a single label prediction per worker class per item, its cost is negligible compared to the cost incurred in the exploit phase, particularly when~$\alpha$ is~small.

Formally, the cost of the exploration phase is $N\sum_{k \in [M]}p_k.$ On the other hand, the cost of the exploitation phase is approximately $\frac{Np_{k^*}}{d(c_{k^*},0.5)}\log\left(\frac{1}{2.4\alpha}\right).$ When $\alpha$ is small, note that the latter term dominates.

$\bullet$ Finally, we comment on the role of the \hide{free} `free' parameter $\gamma \in (0,1/2).$ Lemma~13 of~\cite{jordan} provides a PAC bound on confusion matrix estimates of the following form: the estimates are $\epsilon$-accurate with probability $\geq 1-\delta,$ if $N$ is large enough, where the values of $\epsilon,$ $\delta,$ and $N$ are jointly constrained in terms of the problem parameters. We tie these three quantities feasibly via the parameter~$\gamma.$ Thus, Theorem~\ref{thm:sym} actually specifies a family of performance guarantees for our algorithm. When $\gamma$ is decreased, the upper bound on the probability of mis-identifying the optimal worker class \emph{decreases}, while the threshold $N_0(\gamma)$ beyond which the same (tighter) bound holds \emph{increases}.
%\end{itemize}

%The proof of Theorem~\ref{thm:sym} is provided in Appendix C of Section~\ref{sec:appendix1}. \par

The case of asymmetric confusion matrices admits an analogous treatment; due to space constraints, this is presented in \TRtext{Appendix~\ref{sec:asym_appendix}}\conftext{Appendix~A in~\cite{FullVersion}}. The proposed algorithm for this case, while structurally similar to Sym-IMCW, uses the spectral methods developed by \cite{JMLR:v15:anandkumar14b} and \cite{jordan} for estimating the confusion matrices. A formal description of this algorithm (called Asym-IMCW), along with a rigorous performance guarantee, can be found in \TRtext{Appendix~\ref{sec:asym_appendix}}\conftext{Appendix~A in~\cite{FullVersion}}.

\hide{The analogous Algorithm~\ref{??} for asymmetric confusion matrices is provided in Appendix A of Section~\ref{sec:asym_appendix}. Although the algorithm is similar, we use the spectral methods of \cite{JMLR:v15:anandkumar14b} and \cite{jordan} to estimate the confusion matrices. The performance guarantee employs two parameters $\gamma_a$ and $\gamma_b$ instead of one in the symmetric case and also guarantees that each item is labelled with $\geq 1-\alpha$ accuracy.}

\section{Heuristic Approaches}
\label{sec:heuristics}

While the algorithms presented in Section~\ref{sec:symmetric} and \TRtext{Appendix~\ref{sec:asym_appendix}}\conftext{Appendix~A in~\cite{FullVersion}} admit formal performance guarantees, we present in this section two heuristic approaches that exploit the confusion matrix estimates from the explore phase during the exploit phase. As noted in Remark~\ref{rem:DT_versus_heuristic}, these approaches ignore the uncertainty in the confusion matrix estimates, and therefore do not admit a formal performance guarantee. However, they perform very well in our empirical evaluations. Throughout this section, we consider general (asymmetric) confusion matrices.

{\bf Adaptive stopping time heuristic:} We first present an adaptive stopping time heuristic. It is based on the following information theoretic bound for the hypothetical setting where the confusion matrices are known a priori (proof similar to that of Lemma~\ref{lemma:infoT}).
\begin{lemma}\label{lemma:infoT_full_information}
Consider an item $j$ with (unknown) true label $\ell_j$, and a worker class~$k.$ For $\alpha\in (0,1),$ consider any algorithm that identifies the true label $\ell_j$ with probability at least $1-\alpha,$ using only label predictions on item~$j$ from worker class~$k,$ with prior knowledge of~$\textbf{C}_k.$ Taking $\bar{\ell_j} = 1-\ell_j,$ the number of label predictions~$\tau_k(\ell_j)$ collected by the algorithm satisfies
\begin{equation*}
    \Exp{\tau_k(\ell_j)} \geq \frac{1}{
    d(c_k(\ell_j),1-c_k(\bar{\ell_j}))}\log\left(\frac{1}{2.4\alpha}\right).
\end{equation*}
\end{lemma}
In essence, given a coin whose bias is known to be either $c_k(\ell_j)$ or $1-c_k(\bar{\ell_j}),$ Lemma~\ref{lemma:infoT_full_information} provides a lower bound on the expected number of tosses (queries) needed to identify the underlying bias of the coin correctly with probability~$\geq (1-\alpha).$ As expected, this lower bound is \emph{smaller} than the lower bound in Lemma~\ref{lemma:infoT}, since it assumes that the learner/algorithm has additional information; note that $$d(c_k(\ell_j),1-c_k(\bar{\ell_j})) > d(c_k(\ell_j), 0.5).$$\footnote{While both our information theoretic lower bounds (Lemmas~\ref{lemma:infoT} and~\ref{lemma:infoT_full_information}) are expressed in terms of the confusion matrices, the former assumes the learning agent does not know the confusion matrix, and must therefore deduce the true label purely by identifying the `bias direction' in the label predictions.} It is further possible to devise a Chernoff stopping rule that seeks to asymptotically (as $\alpha \da 0)$ match this lower bound; we refer to this as the $\mathrm{BiasIdentification}$ routine; see Algorithm~\ref{alg:BI}.

\begin{algorithm}[t]
\caption{BiasIdentification}
\begin{algorithmic}[1]
    \State \textbf{Input}: Worker class $k$, Confusion matrix~$\textbf{C}_k$, Item~$j$, Error tolerance $\alpha$
    \State Initialize $t=0,$ $t_1 = 0,$ $Z_0 = 0,$ $Z_1 = 0$ 
    \State Set $\beta(t, \alpha) = \lf{2t}{\alpha}$
    \While{ $t\leq 1$ OR $Z_0 \leq \beta(t, \alpha)$ OR $Z_1 \leq \beta(t, \alpha)$}
    \State $t \la t+1$
    \State Collect a label prediction $z_t\in \{0,1\}$ on item $j$ from worker class $k$
    \State $t_1 \la t_1 + z_t$
    \State Set $\displaystyle Z_1 = \log \frac{(c_k(1))^{t_1} (1-c_k(1))^{t-t_1}}{(1-c_k(0))^{t_1} (c_k(0))^{t-t_1}}$
    \State Set $Z_0 = -Z_1$
    \EndWhile
    \State \textbf{return} $ \mathbf{I}(Z_1 > \beta(t,\alpha))$  
\end{algorithmic}
\label{alg:BI}
\end{algorithm}

The heuristic approach, which we refer to as \emph{Adaptive Bias Identification} (ABI) proceeds as follows. In the explore phase, for each item, collect a single label prediction from each worker class. Use these label predictions to estimate the confusion matrices, as in Asym-IMCW \TRtext{(see Appendix~\ref{sec:asym_appendix})}\conftext{(see Appendix~A in~\cite{FullVersion})}. Next, based on Lemma~\ref{lemma:infoT_full_information}, define the optimal worker class as $\hat{k}_{\mathrm{ABI}}=$ $$\argmin_k p_k \max\left(\frac{1}{d(\hat c_k(1),1-\hat c_k(0)},\frac{1}{d(\hat c_k(0),1-\hat c_k(1)} \right).$$ Finally, in the exploit phase, for each item~$j$, assign the final label to be the output of $\mathrm{BiasIdentification}(\hat{k}_{\mathrm{ABI}}, \boldsymbol{\mathrm{\hat C}}_{\hat{k}_{\mathrm{ABI}}}, j, \alpha).$

{\bf MLE based heuristic:} Next, we propose a non-adaptive heuristic, where we assign the final label to an item via a maximum likelihood estimation (MLE), pretending that the estimated confusion matrix from the explore phase is exactly accurate. The number of label predictions to be collected is further based on an upper bound on the probability that the MLE mis-identifies the true label. To state the heuristic precisely, we need the following result \TRtext{(proof in Appendix~\ref{sec:mle_appendix})}\conftext{(proof in Appendix~E in~\cite{FullVersion})}.
%we construct a maximum likelihood estimate, $\hat{\ell}_j(MLE)$ of the true label of each item based on the acquired predicted labels. The number of workers $t^{M}_\alpha$ is also fixed based on its estimated confusion matrix and $\alpha$. The algorithm is formalized below in Lemma \ref{lemma:mle}
\begin{lemma}\label{lemma:mle}
Assume that the confusion matrices are known. Then given label predictions from~$t^{M}_\alpha$ workers of class~$k$ on an item $j$, the MLE $\hat{\ell}_j$ of $\ell_j$ equals 0 if the fraction of workers predicting~0 exceeds $\theta_k,$ and $\hat{\ell}_j = 1$  otherwise (ties may be broken arbitrarily). Here, the decision boundary~$\theta_k$ is given by  
\begin{equation*}
    \theta_k = \frac{\lf{c_k(1)}{1-c_k(0)}}{\lf{c_k(0)}{1-c_k(1)} + \lf{c_k(1)}{1-c_k(0)}}.
\end{equation*}
The resulting error probability is bounded as follows:
\begin{equation}
\label{eq:prob_error_MLE}
    \prob{\hat{\ell}_j \neq \ell_j} \leq e^{-t^{M}_\alpha d(\theta_k, c_k(0))} 
\end{equation}
\end{lemma}
We note here that $\theta_k$ satisfies $d(\theta_k, c_k(0)) = d(\theta_k, 1 - c_k(1)),$ i.e., $\theta_k$ may be interpreted as a  KL-midpoint between $c_k(0)$ and $(1-c_k(1)).$ Now, to bound the probability of error from above by~$\alpha,$ it follows that $t^{M}_\alpha := \frac{\log(1/\alpha)}{d(\theta_k, c_k(0))}$ predictions suffice. Based on this, the MLE based heuristic acquires $t^{M}_\alpha$ label predictions, but using \emph{estimates} of the confusion matrices from the explore phase. The final label is then assigned using the decision boundary in Lemma~\ref{lemma:mle}.
%\yd{Can we just mention that $\theta$ is a KL-midpoint of.... just as an interesting point, bcz it will help them interpret the lemma better?}

\hide{
The MLE heuristic is fundamentally different from DirectionTest and CBS in the sense that it uses not only the $\hat{k}$ but also $\hat{c}_{\hat{k}}$ from the explore stages of Algorithms \ref{alg:sym} and \ref{alg:asym}. However, an analytical characterization of the performance of this approach is challenging, given the noise in the confusion matrix estimates, and its impact on the actions of the exploit phase. 
}

%OLD DESCRIPTION OF THE CHERNOFF STOPPING RULE UNDER ABI
\hide{
Let $p_1 = c_k(1),$ $p_2 = 1-c_k(0).$ \nh{ $p_i$s are prices?} These are the entries in the row (predicted label) corresponding to 1 in $\textbf{C}_k.$ Suppose that after $n$ draws, we get $n_1$ label predictions saying 1, and $n_0$ saying 0. We stop if either of the following hold:
\begin{align*}
&\log \frac{p_1^{n_1} (1-p_1)^{n_2}}{p_2^{n_1} (1-p_2)^{n_2}} > \log\left(\frac{2n}{\alpha} \right), \\
&\log \frac{p_2^{n_1} (1-p_2)^{n_2}}{p_1^{n_1} (1-p_1)^{n_2}} > \log\left(\frac{2n}{\alpha} \right).
\end{align*}
In the former case, output 1, and in the latter case, output 0. \nh{$n_2 = n_0$? }
}
\section{Case Study} \label{sec:case_study}

In this section, we perform a case study to validate the proposed algorithms. 
%We also propose two natural heuristic variants of the earlier proposed algorithms, which are found to satisfy the accuracy threshold, but incur a lower cost.
%
%We begin by consider the case of asymmetric confusion matrices. 
To perform empirical studies under our model, we need the ground truth confusion matrices of the different crowdsourcing worker classes on a binary classification task, to simulate label predictions. We constructed~5 confusion matrices from the dataset provided by \cite{snow-etal-2008-cheap} on the Recognizing Textual Entailment (RTE) task (originally proposed by \cite{dagan}); details can be found in \TRtext{Appendix~\ref{sec:cs_appendix}}\conftext{Appendix~B in~\cite{FullVersion}}.
\hide{
To consider realistic confusion matrices, we use the dataset provided by \cite{snow-etal-2008-cheap} on the Recognizing Textual Entailment (RTE) task (originally proposed by \cite{dagan}). The dataset contains 8000 labels assigned by 164 unique workers on 108 different items.  We select 5 out of these 164 workers to model our worker classes. The (asymmetric) confusion matrix of each class is computed using the label predictions of the selected workers provided in \cite{snow-etal-2008-cheap}, and the ground truths of the items they labelled. These confusion matrices are unknown to our algorithms and are used only to mimic the response from real world workers on any item $j$. In all our experiments, we have set $\alpha=0.05,$ and we work with these same $5$ worker classes throughout.

\textcolor{blue}{unless i am missing something here, all the data points are not synthetic; only our `data' intervention is the number of worker classes 5. so, i feel that we are somewhere in the real-world and synthetic data settings; hopefully, this is coming out cleanly }
} %end hide
%
%Since our performance guarantees depend on the prices $(p_k,\ k \in M)$ , 
Given these confusion matrices, we consider the following pricing model.
$$\text{Model P1: } p_k = e^{5\times d(\min(c_k(0), c_k(1)), 0.5)}$$
Under P1, prices grow exponentially with `quality.' Table \ref{tab:instances} summarizes the instances we consider; P1-Asym uses the asymmetric confusion matrices as described. P1-Sym is the `symmetrized' version of this instance, where the (symmetric) probability of accurate label prediction is taken as the average of the diagonal entries from the earlier asymmetric matrices. Finally, we set $\alpha = 0.05.$

\hide{Further, we propose two alternative algorithms for the purpose of benchmarking, that differ from the previously proposed algorithms in the exploit phase. The first uses a confidence bound to determine the stopping time (in place of the Chernoff stopping criterion used in DirectionTest). The second is a heuristic that acquires a pre-determined and fixed number of label predictions for all items, using the confusion matrix estimates from the explore phase. We describe these algorithms next, before presenting our empirical results.}

%for our exploit phase algorithm, DirectionTest, we construct two alternate heuristic algorithms, one based on Upper confidence bounds and the other that follows Maximum Likelihood estimation which are described below. The explore phase remains identical, the only change is in the exploit phase. We investigate the effect of these heuristics on the accuracy and the cost of predicting the final label of an item. The motivation behind these is to benchmark our performance against other fixed-confidence or even fixed-budget algorithms for predicting the true label from aggregated responses from a worker class.  \par

\hide{
\paragraph{Confidence Bound based Stopping criterion (CBS):} This is also a stopping time algorithm that uses confidence bounds instead of the Chernoff stopping rule in Algorithm~\ref{alg:dtest}. This is motivated by the fact that for any item~$j,$ $$\left[\hat{c} - \sqrt{\frac{\log(1/\alpha)}{2t}}, \hat{c} + \sqrt{\frac{\log(1/\alpha)}{2t}}\right],$$ where $\hat{c}$  denotes the average prediction~(as also defined in Algorithm~\ref{alg:dtest}), is a confidence interval on $c_{\hat{k}}(\ell_j,1),$ that contains this quantity with probability $\geq 1-\alpha \hide{2\alpha}$ (this follows from the Hoeffding inequality).After $t$ label predictions are collected from the worker class~$\hat{k}$, the CBS algorithm stops if:
\begin{equation}\label{eq:ucb}
    |\hat{c} - 0.5| > \sqrt{\frac{\log(1/\alpha)}{2t}}
\end{equation}
When~\eqref{eq:ucb} holds, the final label $\hat{\ell}_j$ is assigned identical to line 8 of DirectionTest. For this stopping rule, it is easy to see that $\prob{\hat{\ell}_j\neq \ell_j} \leq \alpha.$
}% end hide: Hidden above is the formal description of the CBS approach -- to be moved to an appendix.

\hide{
\begin{align*}
    &\prob{\hat{\ell}_j\neq \ell_j} = \prob{\hat{c}\geq 0.5+ \sqrt{\frac{\log(1/\alpha)}{2t}} \biggr| \ell_j = 0 }\\ &\leq \prob{\hat{c}\geq c_{\hat{k}}(0,1) + \sqrt{\frac{\log(1/\alpha)}{2t}} \biggr|  \ell_j = 0 } \leq \alpha
\end{align*}
}
%Experimentally, we have used plug-in estimators for $\theta_k$ and $t^M_\alpha$ as they are reasonably close to their true values, however weak the performance guarantees might be. Theoretically, another challenge is that carrying over $\hat{c}_k$ to the exploit stage brings the $\delta$ uncertainty with it and to compensate for that, we will require $t^M_{\alpha-\delta}$  instead of $t^M_{\alpha}$ which increases the cost significantly
\hide{
Further, for each of the previously described algorithms, we construct alternate heuristic algorithms, Sym-IMCW-Alt and Asym-IMCW-Alt respectively, where we set $\epsilon=\delta=0$ only in the \emph{exploit} stages of the algorithms. This naturally reduces the number of label predictions, $\hat{s},$ from worker class $\hat{k};$ however,  $\hat{k}$ remains unchanged as that is chosen in the \emph{explore} stage. The motivation behind these heuristics the observation that the concentration bounds in \cite{jordan} on the confusion matrix estimates are \emph{very} loose, and this results in a significant inflation of the number of label predictions requested in the \emph{exploit} stage. The proposed heuristics essentially ignore the confidence bounds on confusion matrix estimates during the exploit stage.
}

\begin{table}
    \centering
    \caption{Summary of instances under pricing model~P1}\label{tab:instances}
    \begin{tabular}{rcccl}
      \toprule % from booktabs package
      \bfseries Instance & $c_{k^*}$ & $p_{k^*}$ & $s_{k^*}$ &   $\Delta_{min}$\\
      \midrule % from booktabs package
      P1-Asym & $(0.88, 0.82)$ & $3.07$ & $30.75$ &$1.48\%$\\
      P1-Sym      &  $0.81$ & $2.95$ & $29.55$ & $1.45\%$\\
      \bottomrule % from booktabs package
    \end{tabular}
\end{table}

\begin{figure*}[t]
    \centering
    \includegraphics[width=0.6\linewidth]{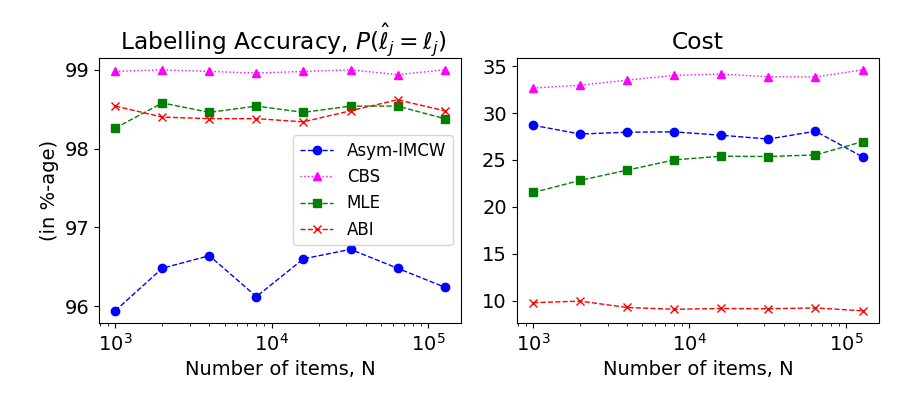}
    \caption{Comparing proposed algorithms (P1-Asym)}
    \label{fig:asym}
\end{figure*}

Due to space constraints, we present only the results corresponding to the instance P1-Asym here; the results corresponding to P1-Sym along with additional comparisons between Asym-IMCW and Sym-IMCW, can be found in \TRtext{Appendix~\ref{sec:cs_appendix}}\conftext{Appendix~B in~\cite{FullVersion}}. The observations are illustrated in Figure~\ref{fig:asym}. Here, CBS stands for a variant of Asym-IMCW, where the Chernoff stopping rule is replaced by one based on confidence bounds \TRtext{(details in Appendix~\ref{sec:cbs_appendix})}\conftext{(details in Appendix~F in~\cite{FullVersion})}. 

$\bullet$~We note that Asym-IMCW does meet the prescribed labelling accuracy. Moreover, the Chernoff stopping rule used in Asym-IMCW outperforms the confidence bounds based stopping criterion from a cost standpoint; the latter approach tends to acquire far more label predictions than needed (this also makes its label assignments more accurate, almost $99\%$ accurate, as compared to the prescribed threshold of $95\%$).

$\bullet$~Interestingly, both heuristics (ABI and the MLE based approach) outperform Asym-IMCW, providing a higher labelling accuracy at a lower or comparable cost. As noted before, this is due to their use of the confusion matrix estimates from the explore phase. This motivates the design of alternative approaches, that perform the tasks of confusion matrix estimation and cost-optimal label assignment jointly--a promising avenue for future work (we remark on this further in Section~\ref{sec:conclusion}).

%\textcolor{blue}{the new heuristic  is actually doing better, i feel. Cost is halved almost, in both Ayym and Sym set ups with almost same accuracy to best one CBS?? -- based on figs in email}
%
%\textcolor{blue}{if above is ok, what could be the reason? yesterday, we discussed a proof in Sym case??}
%We observe in Figure~\ref{fig:asym5_4cols} that under $P_2$, the Algorithm~\ref{alg:asym} selects $\hat{k}$ optimally slightly more often than under $P_1,$ 
%\yd{with both probabilities converging to $1$ in the upper right corners. not sure to say this or not}. 
%This is owing to the fact that under $P_2$, $k^*$ is optimality more because of its low price than its quality. The extra costs in both pricing models are roughly similar because the penalty on selection of a sub-optimal worker class is lower $P_1$ while the probability of selecting it \yd{sub-optimal , in case it is not clear} is higher. For the true label predictions , all the four configurations \yd{to add exact accuracy} . This is an artefact of the Chernoff bounds used to achieve the error tolerance which is tight only asymptotically.

% Contrarily, the extra costs under $P_1$ are slightly lower than in $P_2$    and this is more prominent in \emph{AltAlg2} because they have identical estimates of the confusion matrix and the cost solely depends on choosing the optimal $\hat{k}$. 

%%%%%%%%%%%%%%%%%%%%%%%%%%%%%%%%%%%%%
%%%%%%%%%%%%%%%%%%%%%%%%%%%%%%%%%%%%%
%JK: FOLLOWING PARAS, ON THE SYMMETRIC INSTANCE, AND THE COMPARISON, HAVE TO BE MOVED TO THE APPENDIX ALONGSIDE THE CORRESPONDING PLOTS
\hide{
\paragraph{Symmetric confusion matrices:} Next, we consider the symmetric instances P1-Sym under Sym-IMCW and the two alternative exploit approaches. The results are summarized in Figure~\ref{fig:sym}. The observations and trends are similar to Fig~\ref{fig:asym}, with the MLE heuristic outperforming DirectionTest. Also, the Chernoff stopping rule of  Sym-IMCW outperforms CBS from a cost standpoint.

\paragraph{Comparison:}
Finally we compare our algorithms Asym-IMCW and Sym-IMCW by running the same instance P1-Sym. The results are represented in Figures~\ref{fig:khat} and \ref{fig:sym_in_asym}. Specifically, Figure~\ref{fig:khat} shows the empirical estimates of the probability that the optimal worker class is identified. Note that this appears to be a `harder' task on asymmetric instances, compared to symmetric instances. Moreover, we see that Sym-IMCW outperforms Asym-IMCW on the probability of optimal worker class identification on the symmetric instance. This can be attributed to the more complex machinery in \cite{jordan} for the estimation of asymmetric confusion matrices, as compared to the symmetric one-coin model. However, the labelling accuracy and cost incurred by both algorithms are comparable on the symmetric instance; see Figures~\ref{fig:sym} and~\ref{fig:sym_in_asym}. This is likely due to the small $\Delta_{min}$ in the instance considered here, so that Asym-IMCW incurs a comparable cost in spite of choosing a (slightly) sub-optimal worker class.
} %end hide
%%%%%%%%%%%%%%%%%%%%%%%%%%%%%%%%%%%%%
%%%%%%%%%%%%%%%%%%%%%%%%%%%%%%%%%%%%%

%We observe significant performance gap in Sym-IMCW  and Asym-IMCW when it comes to selecting the optimal worker class as $\hat{k}$ than Asym-IMCW. This can be attributed to the more complex machinery in Algorithm 1 of \cite{jordan} compared to their Algorithm 2. Finally we observe that both perform similarly on the labelling accuracy and cost front. This is likely due to the small $\Delta_{min}$ in the instances

\section{Concluding Remarks}
\label{sec:conclusion}

We model the problem of a requester seeking to perform unsupervised binary classification on a multi-class crowdsourcing platform. The requester seeks to minimize the cost of performing this task, subject to an accuracy constraint. Our proposed algorithms combine flavours of fixed budget as well as fixed confidence MAB algorithms.

The reason we do not use a single-shot MAB style algorithm that combines exploration and exploitation is that the confidence intervals on the confusion matrices (using the spectral machinery developed by Anandkumar et al. (2015) and Zhang et al. (2016)) are \emph{very} loose. Example: a confidence interval of width $\epsilon = 0.1$ is available with probability $\geq 0.9$ only when $N$ exceeds ~$3 \times 10^{33}$ for the instance we consider in our case study. A UCB-style algorithm that uses such confidence intervals would therefore perform very poorly in practice. It is therefore a challenging avenue for future work, to combine the exploration and exploitation aspects of our problem formulation into an algorithm that performs well in practice, and also admits an analytical performance guarantee.

%While the generic asymmetric confusion matrix model is the focus, we also consider the symmetric confusion matrix case; for the latter case, we have stronger cost optimality guarantees. 
%We report our computations/case study on two key performance metrics, the labelling accuracy and the cost incurred. We also introduce two alternative algorithms for benchmarking, one that uses a different stopping rule, and a heuristic based on the confusion matrix estimates from the explore phase or our algorithms. 
%The main takeaways are as follows: First, the Chernoff rule has a smaller query complexity as compared to confidence bound based stopping rules. Second, exploiting the confusion matrix estimations in the exploit phase, which can be interpreted as a joint assignment of labels to items, can improve performance. 

\hide{
\jn{To do:\\
1. Replace the figure in case study with the new one. Use the following legends: Asym-IMCW, MLE, ABI, CBS. Since different tick shapes are used already, I would recommend making the figure grayscale.\\
2. Make sure ABI evaluation is consistent with the description in Section 4.\\
3. Fix appendix references. Make sure there are pointers to the proofs of all stated results.\\
4. Check the formatting of the appendix.
}

\nh{is estimates are denoted by hat, $\hat{c}$ mentioned? }

\nh{a typo: in previous page, right col, `makes' is repeated twice}

\nh{in Algo 3, BiasIdentification, shouldn't the input use $\hat{c}$, hat? as in $\hat{k}_{ABI}$ in argmin below on the same col?}
}
%the probability of correctly identifying the cost-optimal crowd scouring platform and the minimum number of samples needed for the entire model to work. *) parametrisation of asymmetric confusion matricies and its effect on these 2 performance measures *) ... 
 
%Our model has many leads to follow up. For example, in the case of asymmetrical confusion matrices, one can explore the role of hyperparameter pair $(\gamma_a, \gamma_b)$ on various quantities of interest, 
%\nh{to shorten this list?}
%\textbf{has to be changed}
%like the cost minimization of crowdsourcing services; on the performance guarantee of accuracy, $1 - \alpha$; the lower bound on the number of samples needed, $L$; on the high probability of picking the optimal cost worker class $k^*$, $P(\hat{k} = k^*)$; etc. 

%\textbf{i think this is not needed now}Both the models that we proposed are for the fixed budget case, which use specified number of data points to estimate the confusion matrices. An altogether different model would be the one which uses required (adaptive) number of data points that ensure a pre-specified accuracy in identifying the optimal worker class.

% \subsubsection*{References}
\bibliographystyle{IEEEtran}
\bibliography{references}

\onecolumn
\appendices
\section{Asymmetric Confusion Matrices}\label{sec:asym_appendix}

In this section, we consider the general case, where the probability that each worker correctly labels an item can depend on the item's true label. This corresponds to working with asymmetric confusion matrices for each worker class. In this setting, we use the spectral tensor methods developed by \cite{JMLR:v15:anandkumar14b,jordan} to estimate the confusion matrix corresponding to each worker class. 

%Throughout this section, since we need to account for the uncertainty in the estimation of both diagonal entries of each confusion matrix, and the dependence on the true label of an item, we emphasize the choice of the optimal worker class made in ~\eqref{eq:optclass}. Such a choice is pessimal, however it enables us to proceed without depending on the estimated priors as seen in Line 8 of Algorithm \ref{alg:asym}.

Our algorithm for asymmetric confusion matrices, referred to as Asym-IMCW, is presented as Algorithm~\ref{alg:asym}. Asym-IMCW is structurally similar to Sym-IMCW; it also proceeds in two phases---an explore phase for estimating the confusion matrices and thus identifying the optimal worker class with high probability, and an exploit phase (identical to that in Sym-IMCW) for estimating the true item labels with the prescribed accuracy guarantee using label predictions from the estimated optimal worker class.

Next, we provide a brief description of the explore phase of Asym-IMCW. (The exploit phase of Asym-IMCW is identical to that of Sym-IMCW.)

%where the stopping time based \emph{DirectionTest} algorithm is used to acquire labels for each item from workers of the estimated optimal class, and infer the true label with probability at least~$1-\alpha.$

%Quite similar to Algorithm \ref{alg:etc_symmetric_w_rho}, this is also a two -phase algorithm with the \textit{Explore} based on  Algorithm 1 of \cite{jordan}. To collect the labels all $N$ items are sent to each of the $M$ worker classes once, and the collected labels are passed on to Algorithm 1 of \cite{jordan}.

\subsection{Explore Phase of Asym-IMCW}

The \emph{explore} phase of Asym-IMCW is based on Algorithm~1 of \cite{jordan}. Similar to Sym-IMCW, we first obtain a single label prediction for each of the given $N$ items, from all worker classes. 
%These predictions are generated from the latent confusion matrices as defined above and are conditionally independent given the true label of each item. 
Algorithm~1 of \cite{jordan} then partitions the worker classes into three disjoint groups (say~$a, b, $and~$ c$) and estimates group aggregated confusion matrices (up to a column permutation) for all the three groups. This is done by deploying the \emph{robust tensor power method} developed by \cite{JMLR:v15:anandkumar14b} thrice, each time with a different order of~$(a,b,c)$. The correct permutation of the columns is determined by the assumption $c_k(0), c_k(1) > \frac{1}{2}.$ Later, a plug-in estimator for each worker class is used to extract its estimated confusion matrix from that of its group.\par

For the confusion matrix estimators used in Asym-IMCW, Theorem~3 of~\cite{jordan} establishes PAC guarantees, which states that the estimates are~$\epsilon$-accurate with probability at least~$1-\delta,$ for large enough~$N,$ where the permissible tuples of $\epsilon,$ $\delta,$ and the threshold on~$N$ are constrained in terms of the problem parameters. Note that we do not need ~$\epsilon$ and~$\delta$ to be algorithm computable, and use them only to establish our guarantees in Theorem~\ref{thm:asym}. For the sake of convenience, we parameterize these in terms of two floating parameters $\gamma_a, \gamma_b \in (0,1)$ satisfying $\gamma_a + 2\gamma_b \leq 1,$ as shown in the proof of Theorem~\ref{thm:asym}
%$$\epsilon = L^{-\gamma_b}, \quad  \delta = \min\left\{\alpha/2, ~(48+M)\exp\left(1-L^{\gamma_a}\right)\right\}.$$
\hide{Lemma~\ref{lemma:jordan_adapted} in Appendix~\ref{sec:appendix2} shows that these choices do indeed satisfy the PAC bound for large enough~$N.$ }\hide{To prove performance guarantees on Asym-IMCW, we assume that $c^1_k,c^2_k \leq Y < 1$ for all~$k,$ i.e., no worker class is perfect.}

%\begin{equation*} \hat{s}^{(\epsilon, \delta)}_k = \frac{-\log(\alpha-\delta)}{d(\theta(\hat{c}^1_k - \epsilon, \hat{c}^2_k - \epsilon), \hat{c}^1_k - \epsilon)}\end{equation*}

\begin{algorithm}[t]
\caption{Asymmetric - IMCW}
\begin{algorithmic}[1]
    \State \textbf{Input}: prices $\{p_k : k \in [M]\}$; error threshold $\alpha$; number of items $N$; items $\{j : j \in [N]\}$;
    \State $\cdots \cdots \cdots \cdots \cdots \cdots \cdots\cdots\cdots\cdots$ \Comment{Explore Phase}
    \For{$j = 1, \dots, N$}
        \ForAll{$k \in [M]$}
        \State Collect label prediction on item $j$ \hide{from worker $k$}  at price $p_k$
        \EndFor
    \EndFor
    \State Run Algorithm 1 of \cite{jordan} and obtain $\hat{c}_k(0)$ and $\hat{c}_k(1)$  $\forall\ k \in [M]$
    
        \State $\hat{c}_{\hat{k}}(0) \la \max\{\hat{c}_{\hat{k}}(0),\frac{1}{2} \}$; \quad $\hat{c}_{\hat{k}}(1) \la \max\{\hat{c}_{\hat{k}}(1), \frac{1}{2} \}$
    \State Set $\hat{k} = \argmin_{k \in [M] }  \max \{ \frac{p_k}{d(\hat{c}_{\hat{k}}(0), 0.5)}, \frac{p_k}{d(\hat{c}_{\hat{k}}(1), 0.5)}  \}  $
    \State $\cdots \cdots \cdots \cdots \cdots \cdots \cdots\cdots\cdots\cdots$ \Comment{Exploit Phase}
    \ForAll{$j \in [N]$ }       
        \State Assign the \textit{final label}, $\hat{\ell}_j =$  DirectionTest($\hat{k}$, $j$, $\alpha$)
    \EndFor
    
\end{algorithmic}
\label{alg:asym}
\end{algorithm}
\subsection{Performance Guarantee}

We now state our formal performance guarantee for Asym-IMCW. 
\begin{theorem}\label{thm:asym}
% \max(K_1,K_2,K_3, K_4, K_5)
Under the Asym-IMCW algorithm, for each item~$j \in [N],$ $\prob{\hat{\ell}_j \neq \ell_j} < \alpha.$ Moreover, for $\gamma_a,\gamma_b  \in (0,1),$ that satisfy $\gamma_a+2\gamma_b \leq 1,$ if $N \geq N_0(\gamma_a, \gamma_b),$  where $N_0(\gamma_a, \gamma_b)$ is an instance dependent constant, then  
    
\begin{equation*}
    \prob{ \hat{k} \neq k^*} \leq (48+M)\exp\left(1-N^{\gamma_a}\right)
\end{equation*} 

Finally, denoting the number of label predictions acquired for item~$j$ in the exploit phase by $\tau_{\hat{k}},$
\begin{equation*}
     P\left(\limsup_{\alpha\da 0} \frac{\tau_{\hat{k}}}{\log(1/\alpha)}\leq \frac{1}{d(c_{\hat{k}}(\ell_j), 0.5)}\right) = 1.
\end{equation*}
\end{theorem}
The exact expression for $N_0(\gamma_a, \gamma_b)$ is provided in Appendix ~\ref{sec:appendix2}, which also contains the proof of Theorem~\ref{thm:asym}.
\section{Case Study}\label{sec:cs_appendix}

For a representative case study, we wanted to consider a realistic scenario with different worker classes each represented by a unique confusion matrix. We used the dataset provided by \cite{snow-etal-2008-cheap} on the Recognizing Textual Entailment (RTE) task (originally proposed by \cite{dagan}), which had over 8000 labels assigned by 164 unique workers on 108 different items. Post calculating the confusion matrix for each worker, we found that we can choose the following 5 confusion matrices to represent distinct worker classes. Each confusion matrix below is well differentiated from the others to represent a unique class of workers with similar backgrounds and accuracy.

\begin{equation*}
    C_1 = \begin{bmatrix} 0.94 & 0.1\\ 
    0.06 & 0.90
    \end{bmatrix} \quad  C_2 = \begin{bmatrix} 0.77 & 0.13\\ 
    0.23 & 0.87
    \end{bmatrix}  
\end{equation*}
\begin{equation*}
 C_3 = \begin{bmatrix} 0.92 & 0.24\\ 
    0.08 & 0.76
    \end{bmatrix}
 \quad  C_4 = \begin{bmatrix} 0.88 & 0.18\\ 
    0.12 & 0.88
    \end{bmatrix}  
\end{equation*}
\begin{equation*}
    C_5 = \begin{bmatrix} 0.64 & 0.34\\ 
    0.36 & 0.66
    \end{bmatrix}
\end{equation*}

These confusion matrices are unknown to our algorithms and are used only to mimic the response from real world workers on any item $j$. In all our experiments, we have set $\alpha=0.05,$ and work with the same $5$ worker classes for accurate comparison.
\par 
The pricing model P1 ensures that a label from these worker classes is appropriately priced with decreasing marginal utility across worker classes.
\par
\begin{figure}[t]
    \centering
    \includegraphics[width=0.8\linewidth]{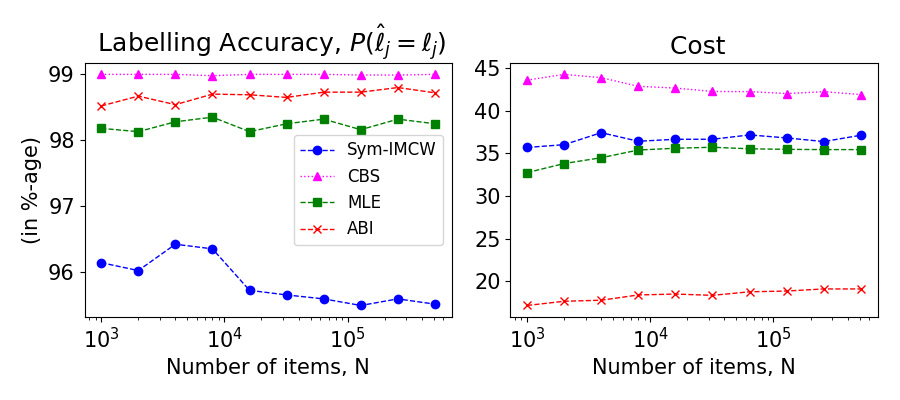}
    \caption{Comparing proposed algorithms on P1-Sym}
    \label{fig:sym}
\end{figure}

\begin{figure}[t]
    \centering
    \includegraphics[width=0.8\linewidth,page=9]{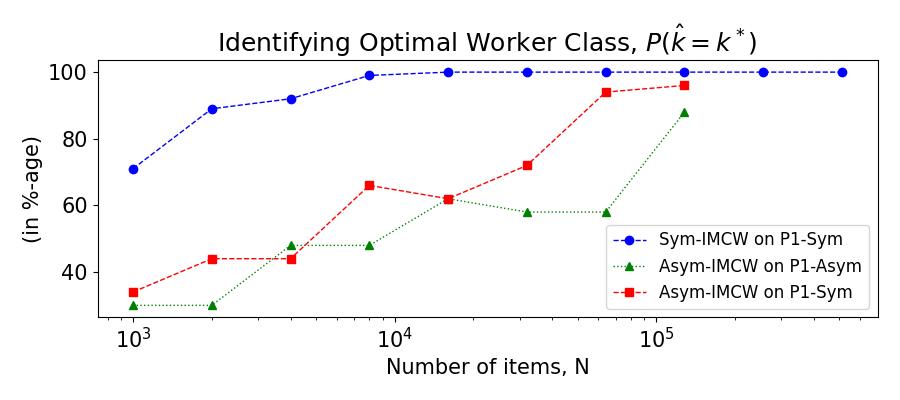}
    \caption{Comparing Asym-IMCW and Sym-IMCW on identification of $k^*$ }
    \label{fig:khat}
\end{figure}

\begin{figure}[t]
    \centering
    \includegraphics[width=0.8\linewidth]{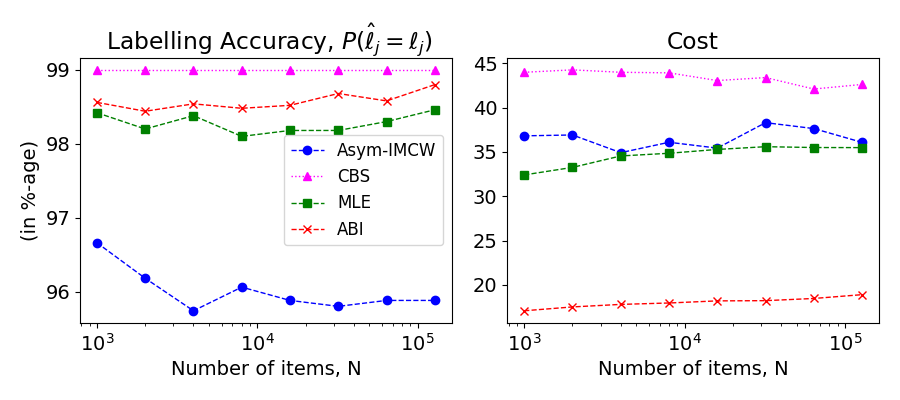}
    \caption{Performance of Asym-IMCW on P1-Sym}
    \label{fig:sym_asymimcw}
\end{figure}

\begin{itemize}
    \item Sym-IMCW performs in a similar fashion as Asym-IMCW qualitatively as seen in Figure~\ref{fig:sym}. The prescribed labelling accuracy of atleast 95\% is met. Sym-IMCW outperforms the CBS heuristic on the cost front, while the other heuristics, ABI and MLE based approaches, provide better accuracy at lower or comparable costs.
    
    \item  We compare our algorithms Sym-IMCW and Asym-IMCW by running them on the same instance P1-Sym. The most interesting comparison of Asym-IMCW and Sym-IMCW is  in their ability to accurately identify the optimal worker class, $k^*$. Figure~\ref{fig:khat} shows the accuracy in identifying worker classes for Asym-IMCW on both instances and Sym-IMCW on P1-Sym. Sym-IMCW clearly outperforms Asym-IMCW here with the latter catching up as number of items increases. This can be attributed to the more complex machinery in \cite{jordan} for the estimation of asymmetric confusion matrices, as compared to the symmetric one-coin model. 
    
    \item The labelling and cost performance of Asym-IMCW on the P1-Sym instance is represented in Figure~\ref{fig:sym_asymimcw}. Asym-IMCW performs very similar to Sym-IMCW on both the labelling and cost standpoint even for smaller number of items. This is likely due to the small $\Delta_{min}$ in the instance considered here (see Table~\ref{tab:instances}), so that Asym-IMCW incurs a comparable cost in spite of choosing a (slightly) sub-optimal worker class.
\end{itemize}


%\pagebreak

\section{Proofs for Symmetric Case}\label{sec:appendix1}

\subsection{Proof of Lemma \ref{lemma:infoT}}
We  structure this problem as a Two-armed bandit with the reward of the first arm $\mu_1$ being either $\mathrm{Bernoulli}(c_k(1))$ or $\mathrm{Bernoulli}(1-c_k(0))$ depending on $\ell_j$. We denote it as $\mathrm{Bernoulli}(c_k(\ell_j, 1))$.  The second arm is taken as deterministic with $\mu_2=\frac{1}{2}$. We thus apply the MAB machinery to decide whether $\mu_1 >\frac{1}{2}$ or $\mu_1 <\frac{1}{2}$ and consequently the estimate the true  label.\\
Theorem 6 of \cite{JMLR:v17:kaufman16a} guarantees that for an identifiable class $\mathcal{M}$ of two-armed bandits and $\nu = (\nu_1, \nu_2) \in \mathcal{M}$ be such that $\mu_1\geq \mu_2$, any algorithm that is $\alpha-$PAC on $\mathcal{M}$ satisfies
\begin{equation*}
    \Exp{\tau} \geq \frac{1}{c_*(\nu)}\lf{1}{2.4\alpha}, 
\end{equation*}
where 
\begin{equation*}
    c_{*}(\nu) := \inf_{(\nu^{'}_1, \nu^{'}_2) \in \mathcal{M} : \mu^{'}_1\leq \mu^{'}_2} \max \{d(\mu_1, \mu^{'}_1), d(\mu_2, \mu^{'}_2)\}
\end{equation*}
In our case, since the second arm is deterministic, the class of bandits, $\mathcal{M}$ is quite restricted and $\mu_2=\frac{1}{2}$ is constant. Hence, the only choice for $\mu^{'}_1$ in $c_{*}(\nu)$ is $\mu^{'}_1 = \frac{1}{2}$ which changes the optimal arm. Hence, for our case, $c_{*}(\nu) = d(\mu_1, \frac{1}{2}) = d(c_k(\ell_j, 1), 0.5) =  d(c_k{\ell_j}, 0.5) $ since $d(x, 0.5) = d(1-x, 0.5)$

\subsection{ Proof of Lemma \ref{lemma:dtest}}
Our complete proof for Lemma \ref{lemma:dtest} is described as follows.
We  structure this problem as a Two-armed bandit with the reward of the first arm $\mu_1$ being either $\mathrm{Bernoulli}(c_k(1))$ or $\mathrm{Bernoulli}(1-c_k(0))$ depending on $\ell_j$. We denote it as $\mathrm{Bernoulli}(c)$, with $c=c_k(\ell_j)$ taking the required value.  The second arm is taken as deterministic with $\mu_2=\frac{1}{2}$. We thus apply the MAB machinery to decide whether $\mu_1 >\frac{1}{2}$ or $\mu_1 <\frac{1}{2}$ and consequently the estimate $\hat{\ell}_j$.\\
Theorem 10 of \cite{pmlr-v49-garivier16a} guarantees that for any sampling strategy, the Chernoff stopping rule with threshold \begin{equation*}
    \beta(t,\alpha) = \log\left(\frac{2t}{\alpha}\right)
\end{equation*}, ensures that the better of the two arms is selected at the end with probability $\geq 1-\alpha$.\\
For the second part,  we use Proposition 13 of \cite{pmlr-v49-garivier16a}. For our set of arms, $w^*_1 = 1$ and $w^*_2 = 0$. Consequently we only sample the first arm , ensuring $N_1(t) = t$ which satisfies all the conditions of Proposition 13 that guarantees for all $\alpha\in(0,1)$,\begin{equation*}
    P_{\mathbf{C}_k}\left(\limsup_{\alpha\rightarrow 0} \frac{\tau_{k}(\ell_j)}{\log(1/\alpha)}\leq \frac{1}{d(c_k(\ell_j), 0.5)}\right) = 1
\end{equation*}   
In other words, the stopping time algorithm achieves the information theoretical lower bound almost surely as $\alpha$ goes down to 0

\subsection{Proof of Theorem \ref{thm:sym}}
The performance guarantee in Theorem \ref{thm:sym} is stated in terms of following instance-dependent constants:
\begin{align*}
    K_1 &= \left( \frac{18}{ \bar{\kappa} \kappa^3_3 } \right)^{1/\gamma}  \\
    K_2 &= \left(\frac{18}{\kappa^3_3 \left( c_{k^*} -  \frac{\log 2 - W_{k^*}}{\log(1/(\log2 - W_{k^*}))} \right)}\right)^{1/\gamma} \\
    K_3 &= \max_{k\in[M]\setminus k^*} \left(\frac{18}{\kappa^3_3 \left( \frac{\log 2 - W_k}{2\log(6/(\log2 - W_k))} - c_k \right)}\right)^{1/\gamma}
\end{align*}
Here, $\gamma$ is a scalar in $(0,1/2),$ $\bar{\kappa} = \frac{1}{M}\sum^{M}_{k=1}(c_k - 0.5),$ $\kappa_3$ is the third largest element in  $\{c_k - 0.5\}^M_{k=1},$ and
\begin{align*}
W^{-1}_k &= \frac{1}{d(c_{k}, 0.5)} - \frac{\Delta_{k}}{2p_{k}} \quad \text{ for } k \in [M]\setminus\{k^*\},\\
W^{-1}_{k^*} &= \frac{1}{d(c_{k^*}, 0.5)} + \frac{\Delta_{min}}{2p_{k^*}}.
\end{align*}
Finally, with $N_0(\gamma) = \max(K_1, K_2, K_3)$ and $N \geq N_0(\gamma)$ as in the statement of Theorem \ref{thm:sym}, our complete proof is described below. 
\par
\begin{proof}
    Running steps (1)-(2) of \textit{Algorithm 2} of \cite{jordan} with $N$ items, returns the estimates $\{\hat{c}_k\}^M_{k=1}$.  Using Lemma \ref{lemma:jordanAlgrho} with $N\geq\left( \frac{18}{\bar{\kappa}\kappa^3_3}\right)^{1/\gamma}$, we obtain 
    \begin{equation*}
       \max_{k \in [M]}\{ \|\hat{c}_k - c_k\| \} \leq \frac{18}{\kappa^3_3 N^{\gamma}} ~ 
    \end{equation*}
    \textit{ w.p } $\geq  1- M^2 \exp\left( -\frac{N^{1-2\gamma}}{2} \right) .$
     Let $G_k:= \{ |\hat{c}_k - c_k | < \epsilon \}$  and let $G = \bigcap G_k $. Observe that under the event $G_k$, for any $\epsilon, \delta > 0$,  it is true that 
    \begin{equation*}
         \frac{1}{d( c_k +  \epsilon, 0.5)} \leq \frac{1}{d(\hat{c}_k, 0.5  )} \leq \frac{1}{d(c_k - \epsilon, 0.5)}
     \end{equation*}
    Setting $\epsilon = \frac{18}{\kappa^3_3 N^{\gamma}}$ and $\delta = M^2 \exp\left( -\frac{N^{1-2\gamma}}{2} \right)$,  we can prove the theorem as follows
    
    \begin{align*}
        &\prob{\hat{k} \neq k^*} \\ &=
        \prob{ \frac{p_{k^*}}{d( \hat{c}_{k^*}, 0.5)}  >  \min_{k \in [M]\setminus\{k^*\} } \frac{p_{k}}{d(\hat{c}_{k},0.5)}}\\
        &=
        \prob{\bigcup_{k \in [M] \setminus k^*  }  \frac{p_{k^*}}{d(\hat{c}_{k^*},0.5)}  >  \frac{p_{k}}{d( \hat{c}_{k},0.5)} } \\ 
        &\leq \prob{G}\prob{\bigcup_{k \in [M] \setminus k^*}  \frac{p_{k^*}}{d( \hat{c}_{k^*},0.5)}  >  \frac{p_{k}}{d( \hat{c}_{k},0.5)} \mid G}  + \prob{G^C} \\ 
         &<  \prob{  \bigcup_{k \in [M] \setminus k^*   }  \frac{p_{k^*}}{d(c_{k^*} - \epsilon,0.5 )} > \frac{p_k}{d( c_k + \epsilon,0.5)} }  + \delta \\
         &= \prob{  \bigcup_{k \in [M] \setminus k^*   }  \frac{p_{k^*}}{d(c_{k^*} - \epsilon,0.5 )} - \frac{p_{k^*}}{d(c_{k^*},0.5)} > \frac{p_k}{d(c_k+\epsilon,0.5)} -\frac{p_{k}}{d(c_{k},0.5)} + \Delta_k } + \delta \\
         &\leq 
          \prob{\frac{p_{k^*}}{d(c_{k^*} - \epsilon,0.5 )} - \frac{p_{k^*}}{d(c_{k^*},0.5)} > \Delta_{min}/2 } +
         \prob{\bigcup_{k \in [M] \setminus k^*   } \frac{p_{k}}{d(c_{k},0.5)} - \frac{p_{k}}{d(c_{k}+\epsilon,0.5)} > \Delta_{k}/2 } + \delta \numberthis \label{eqn:e1} 
    \end{align*}
    %\textcolor{blue}{either $P(G^C)$ above needs a probability as a multiplier or we cna drop this multiplier and have inequality?}
    The  additional requirement on $N$ gives us
    \begin{enumerate}
        \item   For $k=k^*$
        \begin{align*}
            &N \geq  \left(\frac{\kappa^3_3 \left( c_k -  \frac{\log 2 - W_k}{\log(1/(\log2 - W_k))} \right)}{18}\right)^{-1/\gamma}\\
            &\implies c_k - \epsilon \geq \frac{\log 2 - W_k}{\log(1/(\log2 - W_k))} \\ &\geq H^{-1}(\log2-W_k)\\
            &\implies H(c_k - \epsilon) \leq \log2 - W_k ~ \text{(Lemma \ref{lemma:bef})%\textbf{Loose bound}
            } 
            \\
            &\implies d(c_k-\epsilon, 0.5) \geq W_k  \\
            &\implies \frac{p_{k}}{d(c_{k}-\epsilon,0.5 )} - \frac{p_{k}}{d(c_{k},0.5)} \leq \Delta_{min}/2 \numberthis\label{eq:abcd1}
        \end{align*}
        
        \item $\forall k \in [M]\setminus k^*$,
        \begin{align*}
            &N \geq  \left(\frac{\kappa^3_3 \left( \frac{\log 2 - W_k}{2\log(6/(\log2 - W_k))} - c_k  \right)}{18}\right)^{-1/\gamma}\\
            &\implies c_k + \epsilon \leq \frac{\log 2 - W_k}{2\log(6/(\log2 - W_k))}\\ &\leq H^{-1}(\log2 - W_k) \\
            &\implies H(c_k + \epsilon) \geq \log2 - W_k \\ 
            &\implies d(c_k+\epsilon, 0.5) \leq W_k ~ \text{(Lemma \ref{lemma:bef})}
            \\
            &\implies \frac{p_{k}}{d(c_{k},0.5 )} - \frac{p_{k}}{d(c_{k} + \epsilon,0.5)} \leq \Delta_{k}/2 \numberthis\label{eq:abcd2}
        \end{align*}
    \end{enumerate}
    \hide{
    \begin{align*}
        \epsilon = \frac{18}{\rho^3 L^{\gamma}} \leq \min_{k\in[M]} \left(  c_{k} - \frac{1+ \sqrt{1 - e^{-2/ W_k}}}{2} \right) \\ 
        \implies 
        \frac{p_{k}}{d(0.5,c_{k} )} - \frac{p_{k}}{d(0.5,c_{k} + \epsilon)} \leq \Delta_{k}/2 ~ \forall k \in [M]\setminus k^* \\ 
        \text{and, } \frac{p_{k^*}}{d(0.5,c_{k^*} - \epsilon )} - \frac{p_{k^*}}{d(0.5,c_{k^*})} \leq \Delta_{min}/2  \numberthis \label{eq:e4} \\ \text{(By Lemma \ref{lemma:kl_bound})}
        % \text{where, }
        % W_k &= \frac{\Delta_{k}}{2p_{k}} + \frac{1}{d(0.5,c_{k})} ~ \forall k \in [M]\setminus k^* \\ 
        % \text{and, }  W_{k^*} &= \frac{\Delta_{min}}{2p_{k^*}} + \frac{1}{d(0.5,c_{k^*})}
    \end{align*}
    }
    Equation (\ref{eq:abcd1}) and (\ref{eq:abcd2})  ensure that the first term in Equation  (\ref{eqn:e1}) is a zero probability event with the given lower bound on $N$ . And thus we can complete  Equation  (\ref{eqn:e1}) as $ \prob{\hat{k} \neq k^*}  < \delta $ to complete the proof for the theorem

    The proof of the Exploit Stage is provided in Lemma \ref{lemma:dtest}

\end{proof}

\begin{lemma}\label{lemma:jordanAlgrho}
Given $M$ available prices \{$p_k$ : $k\in [M]$ \} with unknown $2\times 2$ symmetric confusion matrices $C_k$ for $k\in [M]$, $L$  different items drawn from a $2$-class distribution and $\rho$ such that $\kappa_3 \geq \rho$, where $\kappa_3$ is the third largest element in  $\{\mid c_k - 0.5 \mid \}^M_{k=1}$, then for any scalar $0<\gamma<0.5$ ,   if the number of items, $L$, satisfies,

\begin{equation*}
    L \geq \left( \frac{ 18}{ \bar{\kappa} \rho^3 } \right)^{1/\gamma} 
\end{equation*}

 ,where $ \bar{\kappa} = \frac{1}{M}\sum^{M}_{k=1}(c_k - 0.5)$ , then the estimates of the confusion matrices returned by Steps(1)-(2) of Algorithm  2 of \cite{jordan} are bounded as
 \begin{equation*}
 \max_{k \in [M]}\{ \|\hat{c}_k - c_k\| \} \leq \frac{18}{\rho^3 L^{\gamma}}      
 \end{equation*}
 with probability $\geq 1-M^2\exp(-\frac{L^{1-2\gamma}}{2})$
\end{lemma} 

\begin{proof}
    The Lemma 13 of \cite{jordan} says that for any scalar, $0<t< \frac{\bar{\kappa} \kappa^3_3}{18}$,
    \begin{equation*}
      \max_{k \in [M]}\{ \|\hat{c}_k - c_k\| \} \leq \frac{18t}{\kappa^3_3}  
    \end{equation*}
    w.p $\geq  1- M^2\exp\left(-\frac{L t^2 }{2 }\right)$
    \\Now  set $t = \frac{1}{L^\gamma}$, we get \begin{equation*}
       \max_{k \in [M]}\{ \|\hat{c}_k - c_k\| \} \leq \frac{18L^{-\gamma}}{\kappa^3_3} \leq \frac{18}{\rho^3 L^{\gamma}}  
    \end{equation*}
     w.p $> 1- M^2\exp\left(-\frac{L^{1-2\gamma}}{2 }\right) $
\end{proof}

\begin{lemma} \label{lemma:kl_bound}
For  any $x > 0.5$, $$d(0.5, x) \geq y \iff   x \geq \frac{1 + \sqrt{1-e^{-2y}}}{2}$$
\end{lemma}
\begin{proof}
    Observe that, $$d(0.5, x) = \frac{-\log(4x(1-x)}{2} > y$$, thus we can say \begin{align*}
        4x(1-x) &\leq e^{-2y} \\
         4x^2 - 4x + e^{-2y} &\geq 0 \\
        x &\geq \frac{1 + \sqrt{1-e^{-2y}}}{2}
    \end{align*}
\end{proof}

\begin{lemma}\label{lemma:bef}
The binary entropy function , $H(x) = -x\log(x) - (1-x)\log(1-x)$ satisfies the following \begin{equation*}
    \frac{x}{2\log(\frac{6}{x})} \leq H^{-1}(x) \leq \frac{x}{\log(\frac{1}{x})}
\end{equation*}
\end{lemma}
\section{Proofs for Asymmetric Case}\label{sec:appendix2}
The guarantee in Theorem~\ref{thm:asym} is stated in terms of the following instance-dependent quantities: 
\begin{align*}
    w_{min} &= \min \{w_0, w_1\},\quad  
    \kappa =  \min_{k \in [M]}  \min_{i \in \{0,1\}}  (2c_{k}(i) - 1),\\
    W &= diag(w_0, w_1), \quad 
    \sigma_L = \min_{a \neq b \in [M]} \norm{\textbf{C}_a W \textbf{C}_b^T},
\end{align*}
\begin{align*}
    K_1 &= \left(\frac{ (72 \times 31 \times 230)^2  2^5 }{w_{min}^2 \sigma_L^{13}}\right)^{1/(1-\gamma_a - 2\gamma_b)}, \\
    K_2 &= \left(  \frac{w_{min} \sigma_L}{72 \kappa}\right)^{1/\gamma_b}, \\
    K_3 &=  \left( c_{k^*}(m_{k^*}) -  \frac{\log 2 - W_{k^*}}{\log(1/(\log2 - W_{k^*}))} \right)^{-1/\gamma_b}, \\
    K_4 &= \max_{k\in[M]\setminus{k^*}}   \left( \frac{\log 2 - W_k}{2\log(6/(\log2 - W_k))} - c_k(m_k)  \right)^{-1/\gamma_b} \\
    K_5 &= \max_{k\in[M]\setminus{k^*}}  \left|{\frac{c_k(0) - c_k(1)}{2}}\right| ^{-1/\gamma_b}.\\
\end{align*}
Here $\gamma_a$ and $\gamma_b$ are scalars in $(0,1)$ that satisfy $\gamma_a + 2\gamma_b \leq 1$, and 
\begin{align*}
    m_k & :=\argmin_{i\in\{0,1\}}\{c_k(i)\} \\
    W^{-1}_k &= \frac{1}{d(c_{k}(m_k), 0.5)} - \frac{\Delta_{k}}{2p_{k}}   ~ \forall k \in [M] \setminus k^* \\
    W^{-1}_{k^*} &= \frac{1}{d(c_{k^*}(m_{k^*}) , 0.5)} + \frac{\Delta_{min}}{2p_{k^*}}
\end{align*}
Finally, with $N_0(\gamma_a, \gamma_b) = \max(K_1, K_2, K_3, K_4, K_5)$ and $N \geq N_0(\gamma_a, \gamma_b)$ as in the statement of Theorem \ref{thm:asym}, our complete proof is described below. 
\par

\begin{proof}
Using Lemma \ref{lemma:jordan_adapted} with $N$ items which satisfy its requirements, we obtain
\begin{equation*}
 \max_{k \in [M]} \norm{\hat{C}_k - C_k}_{\infty} \leq \epsilon =  N^{-\gamma_b} 
 \end{equation*}
  \textit{w.p } $\geq 1 - (48+M)\exp\left(1-N^{\gamma_a}\right)$
 Let $G_k := \{ \norm{\hat{C}_k - C_k}_{\infty} \leq \epsilon \}$ and let $G = \bigcap G_k$.
     Under the event $G_k$ that the estimates are bounded  for  $2$-class asymmetric confusion matrices. Additionally define, $c^m_k = \min\{c_k(0), c_k(1)\} \forall k$ such that \begin{equation*}
         \frac{p_k}{d(c^m_k, 0.5)} = \max \left\{\frac{p_k}{d(c_k(0), 0.5)}, \frac{p_k}{d(c_k(1), 0.5)}\right\}
     \end{equation*}
     Note that introducing the constraint that $\epsilon \leq \left|\frac{c_k(0) - c_k(1)}{2}\right| \forall k \in[M]\setminus k^*$ ensures that \begin{equation*}
         \frac{p_k}{d(\hat{c}^m_k, 0.5)} = \max \{\frac{p_k}{d(\hat{c}^1_k, 0.5)}, \frac{p_k}{d(\hat{c}^2_k, 0.5)}\}
     \end{equation*}
     in other words the estimates $\hat{c}_k$ will be ordered identical to their true values. Now we can work in a similar manner as Theorem \ref{thm:sym} with \begin{equation*}
         \Delta_k = \frac{p_k}{d(c^m_k,0.5)} - \frac{p_{k^*}}{d(c^m_{k^*},0.5)} \forall k \in [M]\setminus k^*
     \end{equation*}
    Setting $\epsilon$  as above and $\delta = (48+M)\exp\left(1-N^{\gamma_a}\right)$,  we can prove part (1) of the theorem as follows
    
    \begin{align*}
        &\prob{\hat{k} \neq k^*} \\ &=
        \prob{ \frac{p_{k^*}}{d( \hat{c}^m_{k^*}, 0.5)}  >  \min_{k \in [M]\setminus\{k^*\} } \frac{p_{k}}{d(\hat{c}^m_{k},0.5)}}\\
         &<  \prob{  \bigcup_{k \in [M] \setminus k^*   }  \frac{p_{k^*}}{d(c^m_{k^*} - \epsilon,0.5 )} > \frac{p_k}{d( c^m_k + \epsilon,0.5)}   }  + \delta \\
         &= \prob{ \bigcup_{k \in [M] \setminus k^*   }  \frac{p_{k^*}}{d(c^m_{k^*} - \epsilon,0.5 )} - \frac{p_{k^*}}{d(c^m_{k^*},0.5)}  > \frac{p_k}{d(c^m_k+\epsilon,0.5)} -\frac{p_{k}}{d(c^m_{k},0.5)} + \Delta_k  } + \delta \\
         &\leq
          \prob{\frac{p_{k^*}}{d(c^m_{k^*} - \epsilon,0.5 )} - \frac{p_{k^*}}{d(c^m_{k^*},0.5)} > \Delta_{min}/2 } +
         \prob{\bigcup_{k \in [M] \setminus k^*   } \frac{p_{k}}{d(c^m_{k},0.5)} - \frac{p_{k}}{d(c^m_{k}+\epsilon,0.5)} > \Delta_{k}/2 }  + \delta
          \numberthis \label{eqn:e_asym1} 
    \end{align*}
    The  additional requirement on $N$ gives us
    \begin{enumerate}
        \item  For $k=k^*$
        \begin{align*}
            &N \geq  \left( c^m_k -  \frac{\log 2 - W_k}{\log(1/(\log2 - W_k))} \right)^{-1/\gamma_b}\\
            &\implies c^m_k - \epsilon \geq \frac{\log 2 - W_k}{\log(1/(\log2 - W_k))}\\
            &\geq H^{-1}(\log2-W_k)\\
            &\implies H(c^m_k - \epsilon) \leq \log2 - W_k ~ \text{(Lemma \ref{lemma:bef})}\\
            &\implies d(c^m_k-\epsilon, 0.5) \geq W_k  \\
            &\implies \frac{p_{k}}{d(c^m_{k}-\epsilon,0.5 )} - \frac{p_{k}}{d(c^m_{k},0.5)} \leq \Delta_{min}/2 \numberthis\label{eq:efgh1}
        \end{align*}
        
        \item $\forall k \in [M]\setminus k^*$,
        \begin{align*}
            &N \geq \left( \frac{\log 2 - W_k}{2\log(6/(\log2 - W_k))} - c^m_k  \right)^{-1/\gamma_b}\\
            &\implies c^m_k + \epsilon \leq \frac{\log 2 - W_k}{2\log(6/(\log2 - W_k))}\\
            &\leq H^{-1}(\log2 - W_k) \\
            &\implies H(c^m_k + \epsilon) \geq \log2 - W_k \\ 
            &\implies d(c^m_k+\epsilon, 0.5) \leq W_k ~ \text{(Lemma \ref{lemma:bef})} \\
            &\implies \frac{p_{k}}{d(c^m_{k},0.5 )} - \frac{p_{k}}{d(c^m_{k} + \epsilon,0.5)} \leq \Delta_{k}/2 \numberthis\label{eq:efgh2}
        \end{align*}
        \item $\forall k \in [M]\setminus k^*$,
        \begin{align*}
            &N \geq \left|\frac{c_k(0)-c_k(1)}{2} \right|^{-1/\gamma_b}\\
            &\implies \epsilon \leq \left|\frac{c_k(0)-c_k(1)}{2}\right| \numberthis\label{eq:efgh3}
        \end{align*}
    \end{enumerate}
    
    Equations (\ref{eq:efgh1}), (\ref{eq:efgh2}) and (\ref{eq:efgh3})  ensure that the first term in Equation  (\ref{eqn:e_asym1}) is a zero probability event with the given lower bound on $N$ . Thus we can complete the proof of the \textit{Explore} stage after Equation \ref{eqn:e_asym1} as $\prob{\hat{k}\neq k^*} \leq \delta$.
     
     The proof of the Exploit Stage is provided in Lemma \ref{lemma:dtest} 
\end{proof}

\begin{lemma} \label{lemma:jordan_adapted}
Given scalars $\gamma_a, \gamma_b >0$ such that $\gamma_a + 2\gamma_b \leq 1$, and $M$ available prices \{$p_i$ : $i\in [M]$ \} with unknown \textbf{$k\times k$} confusion matrices $C_i$ for $i\in [M]$ and $L$  different items drawn from a $k$-class distribution with priors $W = (w_1, \cdots, w_k)$ and 1 observed label on each item from each of the available prices,  then if the number of items satisfy,  $L$
\begin{equation*}
L \geq  \left(\frac{ (72 \times 31 \times 230)^2  k^5 }{w_{min}^2 \sigma_L^{13}}\right)^{1/(1-\gamma_a - 2\gamma_b)}  \text{, and}
L \geq \left(\frac{w_{min} \sigma_L}{36k \kappa}\right)^{1/\gamma_b}   
\end{equation*}
% \frac{ (72 \times 31 \times 230)^2  k^5 }{\epsilon^2 w_{min}^2 \sigma_L^{13}} (1 +  \log((36 + 6k +M)/\delta))
, then the confusion matrices returned by Algorithm  1 of \cite{jordan} are bounded as\begin{equation*}
 \norm{\hat{C}_i - C_i}_{\infty} \leq L^{-\gamma_b} ~ ~ \textit{for all }i\in [M]   
\end{equation*}
 with probability $\geq 1-\delta$ where $delta = (36+6k+M)\exp\left(1-L^{\gamma_a}\right)$  and
\begin{enumerate}
     \item $ \sigma_L = \min_{i \neq j \in [M]} \norm{C_i W C_j^T} $ 
     \item $ \kappa =  \min_{i \in [M]}  \min_{l \in [K]}  \min_{c \in [M]\setminus{l}}  C_{ill} - C_{ilc} $  %\yd{Minimum difference between a diagonal element and a non-diagonal element of the same column}
 \end{enumerate}
\end{lemma}
\begin{proof}
    Setting $\epsilon = L^{-\gamma_b}$ and $\delta = (48+M)\exp\left(1-L^{\gamma_a}\right) $ in Theorem 3 of \cite{jordan}  and considering that we have 2 classes of items, we get the required lower bound on $L$  as 
    \begin{align*}
        L &\geq \frac{ (72 \times 31 \times 230)^2  2^5 }{\epsilon^2 w_{min}^2 \sigma_L^{13}} (1 +  \log((36 + 6k +M)/\delta)) \\ 
        &= \frac{ (72 \times 31 \times 230)^2  2^5 }{L^{-2\gamma_b} w_{min}^2 \sigma_L^{13}} (1 +  L^{\gamma_a}  - 1) \\ 
        &= \frac{ (72 \times 31 \times 230)^2  2^5 }{w_{min}^2 \sigma_L^{13}} (L^{\gamma_a + 2\gamma_b}  ) \\ 
        L &\geq  \left(\frac{ (72 \times 31 \times 230)^2  2^5 }{w_{min}^2 \sigma_L^{13}}\right)^{1/(1-\gamma_a - 2\gamma_b)} \\
    \end{align*}
    The requirement on $\epsilon$ in Theorem 3 of \cite{jordan} is satisfied because
    \begin{align*}
        L &\geq \left(\frac{w_{min} \sigma_L}{72 \kappa}\right)^{1/\gamma_b}    \\ 
        \epsilon =  L^{-\gamma_b} &\leq \frac{72\kappa}{w_{min} \sigma_L}
    \end{align*}
\end{proof}

\section{The MLE Heuristic}\label{sec:mle_appendix}
The MLE based heuristic is based on Lemma~\ref{lemma:mle}. Our proof for Lemma~\ref{lemma:mle} is below
\par
\begin{proof}
Let $Z$ denote the vector of  $t^M_\alpha$  observed labels from as many workers. For simplicity of notation, take the number of labels as $n=t^M_\alpha$ and let $n_0$ and $n_1$ denote the number of labels that are 0 and 1 respectively. Now we consider the likelihood of $Z$ under $\ell_j$, and say that the MLE $\hat{\ell}_j=1$ , when

%\textcolor{blue}{RHS should have $\ell_j = 1$??}
\begin{align*}
    \prob{Z \mid \ell_j =0} &> \prob{Z\mid \ell_j =1} \\
    \text{or, } (c_k(0))^{n_0} (1-c_k(0))^{n_1} &>  (1-c_k(1))^{n_0} (c_k(1))^{n_1} \\
    \text{or, } \left(\frac{c_k(0)}{1-c_k(0)}\right)^{n_0} (1-c_k(0))^{n} &>  \left(\frac{1-c_k(1)}{c_k(1)}\right)^{n_0} (c_k(1))^{n} \\
    \text{or, } \frac{n_0}{n} &> \frac{\lf{c_k(1)}{1-c_k(0)}}{\lf{c_k(0)}{1-c_k(1)} + \lf{c_k(1)}{1-c_k(0)}} \eqqcolon \theta_k \numberthis \label{eqn:theta}
 \end{align*} 
% \textcolor{blue}{can use $\eqqcolon$ before $\theta_k$??}
 
In other words, if the fraction of workers predicting~0 exceeds $\theta_k,$  we assign $\hat{\ell}_j=0$ as stated in the lemma.
%.... \textcolor{blue}{can mask this line}
In order to bound the error probability observe that by Chernoff bound for binomial random variables, we have
$$
\prob{\hat{\ell}_j=1\mid \ell_j=0} \leq \prob{n_0 \leq n\theta_k \mid \ell_j=0 } \leq e^{-n d(\theta_k, c_k(0))} 
$$
as here $n_0 \sim Binomial(n,c_k(0)) $ under $\ell_j=0$. Similarly under $\ell_j=1$, $n_0 \sim Binomial(n,1-c_k(1)) $ 
$$
\prob{\hat{\ell}_j=0\mid \ell_j=1} \leq \prob{n_0 \geq n\theta_k \mid \ell_j=1 } \leq e^{-n d(\theta_k, 1-c_k(1))} 
$$
Finally we show that $d(\theta_k, c_k(0)) = d(\theta_k, 1-c_k(1))$ and complete our proof of the error bound in Eq \eqref{eq:prob_error_MLE} (reproduced below).
\begin{equation*}
    \prob{\hat{\ell}_j \neq \ell_j} \leq e^{-t^{M}_\alpha d(\theta_k, c_k(0))} 
\end{equation*}

Observe that, 
\begin{align*}
&d(\theta_k, c_k(0)) - d(\theta_k, 1-c_k(1)) \\
&= \theta_k\lf{1-c_k(1)}{c_k(0)} + (1-\theta_k)\lf{c_k(1)}{1-c_k(0)}\\
&= \frac{\lf{c_k(1)}{1-c_k(0)}\lf{1-c_k(1)}{c_k(0)} + \lf{c_k(0)}{1-c_k(1)}\lf{c_k(1)}{1-c_k(0)} }{\lf{c_k(0)}{1-c_k(1)} + \lf{c_k(1)}{1-c_k(0)}} ~~ \text{(Substituting $\theta_k$ from Eq~\eqref{eqn:theta})}\\
&= 0
\end{align*}
This completes the proof.
\end{proof}

\section{The CBS Heuristic}\label{sec:cbs_appendix}
In this section, we formally state the \textbf{Confidence Bound based Stopping criterion (CBS)}. \par 
 CBS is also a stopping time algorithm that uses confidence bounds instead of the Chernoff stopping rule in Algorithm~\ref{alg:dtest}. This is motivated by the fact that for any item~$j,$ $$\left[\hat{c} - \sqrt{\frac{\log(1/\alpha)}{2t}}, ~~  \hat{c} + \sqrt{\frac{\log(1/\alpha)}{2t}}\right],$$ where $\hat{c}$  denotes the average prediction~(as also defined in Algorithm~\ref{alg:dtest}), is a confidence interval on $c_{\hat{k}}(\ell_j,1),$ that contains this quantity with probability $\geq 1-\alpha \hide{2\alpha}$ (this follows from the Hoeffding inequality). After $t$ label predictions are collected from the worker class~$\hat{k}$, the CBS algorithm stops if:
\begin{equation}\label{eq:ucb}
    |\hat{c} - 0.5| > \sqrt{\frac{\log(1/\alpha)}{2t}}
\end{equation}
When~\eqref{eq:ucb} holds, the final label $\hat{\ell}_j$ is assigned identical to line 8 of DirectionTest. For this stopping rule, it is easy to see that $\prob{\hat{\ell}_j\neq \ell_j} \leq \alpha.$

\end{document}